\documentclass[preprint,12pt]{elsarticle}

\usepackage{lineno,hyperref}
\usepackage{natbib}
\usepackage{geometry}
\usepackage{fleqn}
\usepackage{graphicx}
\usepackage{newtxtext,newtxmath}
\usepackage{hyperref}
\usepackage{booktabs}
\usepackage{bm}
\usepackage{amsmath}
\usepackage{enumitem}
\usepackage{esvect}
\usepackage{subcaption}
\usepackage{soul}
\usepackage{upgreek}
\modulolinenumbers[5]

\usepackage{caption}
\usepackage[table]{xcolor}
\usepackage{siunitx}

\journal{Elsevier}

\begin{document}

\begin{frontmatter}

\title{\large{Physics-Based Hybrid Machine Learning for Critical Heat Flux Prediction with Uncertainty Quantification}}

\author[NCSU]{Aidan Furlong\corref{mycorrespondingauthor}}
\cortext[mycorrespondingauthor]{Corresponding author}
\ead{ajfurlon@ncsu.edu}

\author[UTK]{Xingang Zhao}

\author[ORNL]{Robert K. Salko}

\author[NCSU]{Xu Wu}

\address[NCSU]{Department of Nuclear Engineering, North Carolina State University,    \\ 
	Burlington Engineering Laboratories, 2500 Stinson Dr., Raleigh, NC 27695 \\}

\address[UTK]{Department of Nuclear Engineering, University of Tennessee, Knoxville, \\ Zeanah Engineering Complex, 863 Neyland Dr., Knoxville, TN 37916}

\address[ORNL]{Nuclear Energy and Fuel Cycle Division, Oak Ridge National Laboratory, \\1 Bethel Valley Rd., Oak Ridge, TN 37830}

\begin{abstract}
Critical heat flux is a key quantity in boiling system modeling due to its impact on heat transfer and component temperature and performance. This study investigates the development and validation of an uncertainty-aware hybrid modeling approach that combines machine learning with physics-based models in the prediction of critical heat flux in nuclear reactors for cases of dryout. Two empirical correlations, Biasi and Bowring, were employed with three machine learning uncertainty quantification techniques: deep neural network ensembles, Bayesian neural networks, and deep Gaussian processes. A pure machine learning model without a base model served as a baseline for comparison. This study examines the performance and uncertainty of the models under both plentiful and limited training data scenarios using parity plots, uncertainty distributions, and calibration curves.

The results indicate that the Biasi hybrid deep neural network ensemble achieved the most favorable performance (with a mean absolute relative error of 1.846\% and stable uncertainty estimates), particularly in the plentiful data scenario. The Bayesian neural network models showed slightly higher error and uncertainty but superior calibration. By contrast, deep Gaussian process models underperformed by most metrics. All hybrid models outperformed pure machine learning configurations, demonstrating resistance against data scarcity.

\end{abstract}


\end{frontmatter}


{\renewcommand\thefootnote{}\footnotetext{This manuscript has been authored by UT-Battelle LLC, under contract DE-AC05-00OR22725 with the US Department of Energy (DOE). The US government retains and the publisher, by accepting the article for publication, acknowledges that the US government retains a nonexclusive, paid-up, irrevocable, worldwide license to publish or reproduce the published form of this manuscript, or allow others to do so, for US government purposes. DOE will provide public access to these results of federally sponsored research in accordance with the DOE Public Access Plan (http://energy.gov/downloads/doe-public-access-plan).}}

\section{Introduction}
 
Critical heat flux (CHF) in a boiling system occurs when a boiling regime changes from nucleate boiling to transition boiling; this change is marked by a sudden decrease in the ability to transfer heat in a two-phase flow system. The physical presentation of CHF can generally be categorized into two mechanisms: departure from nucleate boiling (DNB) and dryout (DO). DNB occurs when steaming leads to a sustained insulating blanket of vapor between the heated surface and the liquid phase, inhibiting heat transfer efficiency. This occurs in low-quality flows, in which the liquid phase still predominantly occupies the flow channel cross section. DO is a phenomenon in high-quality flows, in which there is a gas/entrained droplet core and a liquid film along the walls of the channel. DO is indicated when the liquid film thickness reaches zero. In the case of conventional heat exchangers and boilers, DO can create large inefficiencies and cause tubing to rupture. In a nuclear system, the consequences can be more catastrophic: fuel cladding failure and the potential for a fuel melt. Therefore, CHF is a key safety-related quantity that requires rigorous consideration during the thermal-hydraulic analysis of a nuclear system.

Much work has been performed over the past 80 years in studying CHF. The goal has been to reliably predict its occurrence in a variety of operating conditions in various geometries. Most of these attempts have been in the form of tube experiments, in which a moving fluid at a specified set of thermal-hydraulic conditions is heated until CHF is reached. Several empirical correlations---including the Biasi, Bowring, CISE-4, and W-3 correlations, among others---have been developed from the resulting data~\cite{todreas2021nuclear}. Each of these correlations has a unique range of validity, and some are specialized for bundles or DNB/DO specifically. The 2006 Groeneveld look-up table (LUT), which is widely used throughout the nuclear industry, was also developed by leveraging a compilation of data from 59 experiments \cite{groeneveld2019critical}. This combined dataset, which was used to create the Groeneveld LUT, consists of nearly 25,000 entries that describe the experimental conditions for uniformly heated vertical tubes using seven input variables and the resulting CHF values for cases of both DNB and DO. Although the empirical correlations and the LUT are effective at computing CHF values for a variety of different input combinations, they still have substantial deviations from experimental measurements in various regions of the operational space \cite{groeneveld20072006}. As a result of this and other factors, work has not ceased in attempting to find a more accurate method of predicting CHF.

Some of the most recent developments in this space have been in the use of data-driven machine learning (ML) techniques, such as deep neural networks (DNNs), convolutional neural networks (CNNs), conditional variational autoencoders (CVAEs), support vector machines (SVMs), and random forests (RFs) \cite{jiang2013combination}\cite{kim2021prediction}\cite{zubair2022critical}\cite{helmryd2024investigation}\cite{KHALID2024125441}\cite{alsafadi2024predicting}\cite{qi2025machine}. These approaches rely on fitting parameters to training data and do not possess any knowledge about the physical world or problem. The primary benefits of these approaches are their ability to find relationships that may not be readily apparent in the data---often leading to a higher overall accuracy in comparison with traditional methods---as well as their inexpensive nature and quick performance post-training. In 2022, under the guidance of the Organisation for Economic Co-operation and Development (OECD) Nuclear Energy Agency (NEA), the Task Force on Artificial Intelligence and Machine Learning for Scientific Computing in Nuclear Engineering was formed with the objective of creating an ML benchmark for CHF prediction \cite{lecorre2023benchmark}. Phase one of this project focused on feature analysis and the training and evaluation of ML regression models using the following parameters (or some of them) as inputs: tube diameter (\textit{D}), heated length (\textit{L}), pressure (\textit{P}), mass flux (\textit{G}), and equilibrium quality ($x_\mathrm{e}$).

Although these surrogate model approaches show greatly improved accuracy in comparison with the Groeneveld LUT, one major limitation is in the interpretability of the models themselves. These data-driven methods are often described as ``black boxes''. Because their complexity increases as the number of fitting parameters scales up; these fitting parameters are often on the order of thousands or even millions, depending on the problem and ML technique. Highly sensitive applications, such as nuclear analysis, require the development of a framework that emphasizes explainability and reliability in the models' predictions.

Data scarcity is another substantial concern in attempting to train data-driven models using the limited datasets available when working with experimental data. Traditional ML approaches, such as DNNs, require a relatively large amount of high-quality data (typically on the order of hundreds of thousands of data entries) to achieve acceptable performance. Providing inadequate amounts of data or entries with a high degree of noise can significantly degrade model performance to the point of a complete breakdown of the training process. One attempt \cite{rohatgi2022machine} to solve this concern considered DNB in $5\times5$ bundle data from the pressurized water reactor subchannel and bundle test benchmark (PSBT). This study focused on using generative adversarial networks to augment a 76-point dataset with synthetic values to be later used to train a DNN. Another study \cite{alsafadi2023deep} also attempted to use a generative approach to mitigate this problem in the prediction of void fractions in a boiling water reactor (BWR) bundle benchmark.

Although these approaches have been shown to be effective at mitigating the scarcity concern, they still lack explainability or a guarantee that completely nonphysical results will not be produced. A recently proposed method not based on data augmentation \cite{zhao2020prediction} works to leverage knowledge of the physical world by using a ``base model'' in combination with a data-driven ML model. This hybrid approach first uses an established model, such as an empirical thermal-hydraulic correlation, to compute an estimate for a target output. The estimate value is then corrected with an ML model trained to predict the residual between those outputs with known experimental values. This correction will compensate for the bias and undiscovered mismatch between the domain knowledge-based model and actual observation. This arrangement is classified as a parallel ``gray-box'' approach \cite{thompson1994modeling}, which is easier to interpret \cite{psichogios1992hybrid} because the bulk of physical knowledge is provided by the base model, reducing the amount of inferred knowledge required by the ML component. Structuring the model in this manner offers a larger degree of interpretability, performance benefits in comparison with stand-alone ML methods, and built-in resistance to the deleterious effects of data limitations.

Other studies have examined the use of these hybrid models. One such study compares the use of DNNs, CNNs, and RFs as the ML technique paired with the Biasi correlation, the Groeneveld LUT, or the Zuber correlation as the base model \cite{mao2024uncertainty}; this study used \textit{D}, \textit{L}, \textit{P}, \textit{G}, $x_\mathrm{e}$, $T_{\mathrm{inlet}}$, and $\Delta h_{\mathrm{sub}}$ as input parameters. Uncertainty quantification (UQ) was performed via input perturbation to capture data uncertainty but not uncertainty resulting from other sources, such as randomness in the training process, model architecture, or extrapolation, among others. The RF model using the LUT as the base model was shown to have the smallest relative error. Another study \cite{khalid2023comparison} provides a comparison between the stand-alone LUT and the artificial neural network, SVM, and RF hybrids, each of which used the LUT as the base model. This study concluded that hybrid models exhibit greater accuracy in every case in comparison with the stand-alone LUT and nonhybrid ML variants. The specific location of CHF occurrence, in addition to its magnitude, has also been used as a prediction target in rectangular channels \cite{NIU2024125042}.

In a follow-up study \cite{furlong2024hybrid} to Zhao et al.~\cite{zhao2020prediction}, this hybrid approach implemented the Biasi and Bowring CHF empirical correlations \cite{todreas2021nuclear} as prior models and assessed the performance of these models against a purely data-driven DNN in a series of training set sizes. In all eight of the training size cases, the hybrid models outperformed the pure DNN, and the maximal improvement for the most data-scarce case. In this case, in which the training set had only nine points, the pure DNN's performance decreased to 48.25\% relative error, but both hybrid models still outperformed the stand-alone empirical correlations with relative errors below 6.40\%.

This study builds on work previously performed using hybrid knowledge-based models in the prediction of CHF, specifically for cases of DO. This study is focused on new contributions in three key areas: further improving the performance of the hybrid models via the exploration of various ML approaches, quantifying these models' uncertainties, and assessing the quality of the quantified uncertainty estimates. Three different ML methods are additionally considered herein for their abilities to quantify uncertainty: DNN ensembles, Bayesian neural networks (BNNs), and deep Gaussian processes (DGPs).

This manuscript is organized as follows. Section \ref{sec:background} provides an overview of the theory behind CHF, data-driven ML, and the UQ of ML models. Section \ref{sec:methods} describes the overarching hybrid framework, data handling, and each of the three ML methods used in this study. Section \ref{sec:results} presents the results, organized with respect to each of the ML approaches, as well as a direct comparison of the ML approaches. Section \ref{sec:conclusion} summarizes the key findings and discusses future directions for research and production.

\section{Background} \label{sec:background}

\subsection{Critical Heat Flux}

Exceeding the CHF can result in changes to boiling that present themselves differently in BWRs and pressurized water reactors. In BWRs, boiling is a part of normal operation and is the primary form of heat transfer from the fuel bundles. The boiling column develops from a liquid state through a series of flow regimes: bubbly, slug, and then annular, which consists of a gas core containing entrained liquid droplets. As heat fluxes increase, the film layer on the heated surface decreases in depth. Once CHF is reached, DO can occur; this means that the liquid film has ``dried out,'' leaving an uncovered surface. DO causes a significant rise in temperature on the heated surface, and this temperature rise can cause fuel damage. Because of these potential consequences, a safety-related parameter known as the minimum critical power ratio is implemented in BWRs to ensure adequate margin to DO.

In this study, the DO state is considered exclusively to ensure that the data used are from the same underlying CHF condition. The public CHF dataset contains both DNB and DO experiments, which are separable by establishing a constraint on equilibrium quality ($x_\mathrm{e}$). Instances of DO have far larger equilibrium qualities in comparison with those of DNB, so a conservative threshold was set at $0.2$ \cite{jin2021unified}.

The NRC pubic CHF database, used to construct the 2006 Groeneveld LUT, formed the source dataset for this investigation.  
This compilation contains both DNB and DO experiments, which can effectively be separated based on equilibrium quality ($x_e$).
In this study, the DO state is considered exclusively to ensure that the data used is from the same underlying CHF condition.
As an initial filter on the source dataset, a conservative equilibrium quality threshold of above 0.2 was set, since DO instances have far larger qualities compared to DNB \cite{jin2021unified}.

Outside the Groeneveld LUT, two widely accepted empirical correlations to predict CHF are the Biasi and Bowring correlations \cite{todreas2021nuclear}. Both correlations have constraints in terms of the regions of their validity within the input space. Both correlations have input ranges where they are valid, so a common range for each input parameter is necessary to find a space where both correlations are valid. These ranges were determined, and they are listed in Table \ref{tab:filter_constraints}. All entries in the DO dataset that fell outside these ranges were removed to ensure correlation validity. The two correlations were then applied to this filtered dataset using the heat balance method (HBM)~\cite{hejzlar1996consideration} and compared against known values to assess the correlations' performance in the 6 key metrics displayed in Table \ref{tab:base_correlation_performance}. These values serve as a baseline that this study can be compared against.

\begin{table}[ht!]
    \centering
    \caption{Dataset filtering criteria for adherence to both the Biasi correlation and the Bowring correlation}
    \label{tab:filter_constraints}
    \begin{tabular}{lc}
        \toprule
        Parameter & Criterion \\ \hline
        $D\; (\si{\meter})$ & \numrange{0.003}{0.0375} \\ \hline
        $L\; (\si{\meter})$ & \numrange{0.20}{3.70} \\ \hline
        $P\; (\si{\mega\pascal})$ & \numrange{0.27}{14.0} \\ \hline
        $G\; (\si{\kilo\gram\per\square\meter\per\second})$ & \numrange{136}{6000} \\ \hline
        $\mathrm{Outlet}\; x_\mathrm{e}$ & $\geq 0.2$ \\
        \bottomrule
    \end{tabular}
\end{table}

\begin{table}[ht!]
    \centering
    \caption{Performance metrics of the Biasi and Bowring correlations using the DO dataset after filtering with the common constraints}
    \label{tab:base_correlation_performance}
    \begin{tabular}{lcc}
        \toprule
        Parameter & Biasi & Bowring \\ \hline
        $\upmu_\text{error}$ & $6.935\%$ & $6.778\%$  \\ \hline
        $\text{Max}_{\text{error}}$ & $171.8\%$ & $626.4\%$ \\ \hline
        $\text{Std}_{\text{error}}$ & $9.470\%$ & $14.22\%$ \\ \hline
        $F_{\text{error}}>10\%$ & $19.67\%$ & $18.65\%$ \\ \hline
        $R^2$ & $0.9746$ & $0.9703$ \\
        \bottomrule
    \end{tabular}
\end{table}

In Table \ref{tab:base_correlation_performance}, All error subscripts indicate absolute relative error, $F_{\text{error}}>10\%$ indicates the fraction of points with absolute relative errors above 10\%, and $R^2$ is the coefficient of determination.
\subsection{UQ of Data-Driven Machine Learning}

Within modeling systems, two forms of uncertainty exist: aleatoric uncertainty and epistemic uncertainty. Aleatoric uncertainty arises from the inherent randomness present within data, from aspects such as measurement error, or from the intrinsic variability of the system being observed. This form of uncertainty is irreducible, even when more data are provided. Conversely, epistemic uncertainty represents the limitations of the modeling approach in attempting to capture the underlying target distribution. Epistemic uncertainty can arise from various sources, such as parameter uncertainties, differences in model architectures, and limitations in the training data. As an example, a model trained on a small or unrepresentative dataset cannot adequately learn true patterns and relationships, which results in higher epistemic uncertainty. This form of uncertainty is reducible with better knowledge of the problem, larger datasets, better model architectures, and an optimized modeling technique.

In conventional ML approaches (e.g., DNNs), parameter uncertainty is a key epistemic consideration. Parameter uncertainty specifically arises from the randomness introduced by the stochastic nature of gradient descent and other optimization processes. One example is the initialization of weights in a DNN; initialization will affect the parameter space traversed during training, leading to different optimized weights and predictions in comparison with those of a model of a different initialization. Several techniques, such as dropout, Bayesian approaches, and ensembling, have been developed to mitigate and understand this uncertainty by averaging unique predictions or by representing weights as distributions rather than as single values. Understanding and managing these uncertainties can improve the performance and reliability of ML models. The three methods applied in this study with UQ capabilities are DNN ensembles, BNNs, and DGPs. Detailed descriptions are provided in Sections \ref{sec:methods_ensemble}, \ref{sec:methods_bnn}, and \ref{sec:methods_dgp}.

\subsection{Quality of Uncertainty Estimates} \label{subsec:uq_quality}

An often overlooked aspect of UQ is verifying that the uncertainty estimates are indeed useful. If these estimates are not of high quality, then they can cause misled decision-making processes and degrade the overall purpose of performing UQ. Three pieces of evidence can help assess uncertainty estimate quality: calibration curves, the magnitudes of the estimates, and their spread/distribution \cite{tran2020methods}. 

In the case of UQ, calibration curves are a graphical approach for assessing the \textit{calibration} of a model's uncertainty estimates by comparing the observed cumulative distribution of standard deviation--normalized residuals with the theoretical cumulative distribution of a standard normal distribution \cite{tran2020methods}. This approach is different than the typical reliability diagram, which focuses on the calibration of predictions themselves rather than focusing on reported uncertainties. A model with well-calibrated uncertainty estimates will produce normalized residuals that closely follow a standard normal distribution, resulting in a curve that aligns with the identity line. Analyzing the calibration curve can help identify systematic overconfidence or underconfidence in the model's uncertainty estimates. The right-hand side of the curve falling below the ideal calibration line indicates that uncertainty estimates are too low, signifying overconfidence. Conversely, the right-hand side of the curve falling above the identity line indicates that uncertainty estimates are too high, signifying underconfidence.

The magnitudes of uncertainty estimates in regression correspond to the model’s confidence in its predictions. Uncertainty estimates are often quantified by the distribution of standard deviations
from a set of predictions when given an identical input set (interpreted as samples from an
underlying distribution). If the magnitudes of these estimates are too large, then the models may lack sufficient data or be overly conservative. Overconfident predictions
may also occur with small magnitudes, which can indicate insufficient representation of
model uncertainties.

In the consideration of the distribution of the prediction ``samples,'' the shape and spread of the uncertainty estimates across data points are useful for determining how a model captures variability. In regression, high-quality UQ provides varying uncertainty estimates that reflect different degrees of confidence based on the input features. This spread indicates how the model differentiates between high and low uncertainty regions, which can reflect regions of higher complexity or in extrapolation outside of the training space. Uncertainty estimates that remain constant over all input regions are neither effective nor useful.

\section{Methods} \label{sec:methods}

The basis of the proposed hybrid approach is the prediction of residuals calculated from a base model and experimental data values. The input parameters for this problem were first selected: tube diameter (\textit{D}), heated length (\textit{L}), pressure (\textit{P}), mass flux (\textit{G}), and inlet subcooling ($\Delta h_{\mathrm{sub}}$). The two other parameters in the original CHF dataset---inlet temperature and outlet equilibrium quality---are redundant and can be computed using the other input parameters. Therefore, these parameters may be readily excluded from this input parameter set. A base model can now be selected to provide a ``first-guess'' output value, which will later be corrected by the ML component. In theory, the base model can be any model that provides relevant information that links the inputs to the output value. In the case of this study, the base model corresponds to either the Biasi or Bowring correlations. Other correlations/models, such as the three-field model or even an LUT, could also be appropriate. The Groeneveld LUT was eliminated as an option for this experiment because it was constructed using the same data that were used to test and train the ML model. The use of this LUT would directly leak information during the testing phase because the holdout dataset would be polluted with values that the model could have potentially ``seen'' during training. This issue could lead to inflated performance scores and prevent an accurate assessment of the trained model's ability to generalize to new data. 

\begin{figure}[ht!]
    \centering
    \includegraphics[width=0.8\linewidth]{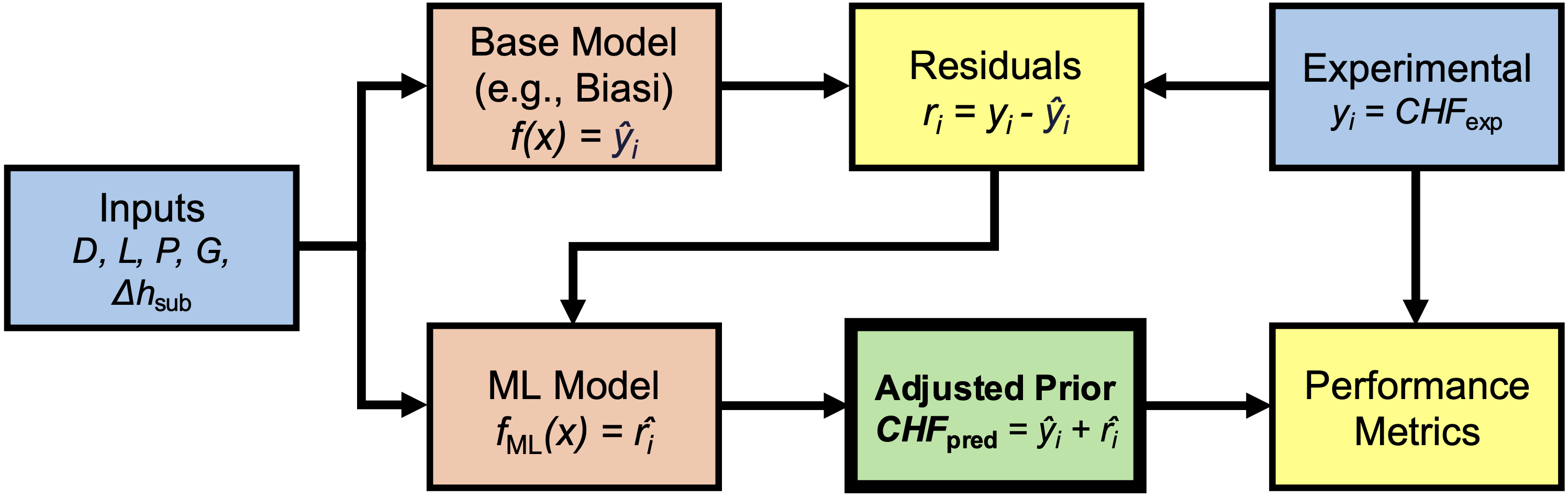}
    \caption{Flowchart of the hybrid ML approach in the \textit{training} configuration. Blue indicates data, pink indicates models, yellow corresponds to ancillary calculations, and green signifies the final output. In the \textit{testing} configuration, the residuals are unknown, so it is entirely up to ML model to adjust the base model’s output solely based on the input vector.}
    \label{fig:flowchart_hybrid}
\end{figure}

Figure~\ref{fig:flowchart_hybrid} provides an overview of the hybrid model training workflow. Once the base model has been selected, the inputs of a given data entry $i$ are fed into the model to compute $\hat{y}_i$, a traditional estimate of CHF. The difference is then taken between this value and a known experimental value $y_i$ to compute the residual ($r_i$), which the ML model will be trained to predict. The data-driven model is then trained using the same set of inputs, producing a predicted output ($\hat{r}_i$). The final prediction ($\Tilde{y}_i$) is given when the predicted residual is added to the base model's output to essentially provide an adjustment to the base model's first estimate, ideally closing the gap between calculated and observed values. This process is the training configuration of this approach. During actual implementation, the experimental data, and therefore the true residuals, are not known; the CHF predictions consist of only the base models' outputs combined with the ML models' predicted residuals. Although the ML component of this method predicts a different quantity than the pure ML method does (i.e., a correlation-experiment residual vs. a direct CHF value), the true output quantity of the two methods remains the same. The analysis and UQ performed in this study focused on this final CHF quantity rather than the hybrids' residuals themselves.

Three ML UQ techniques were considered in this study: DNN ensembles, BNNs, and DGPs. For each of these techniques, three choices of base models were compared: no base model, the Biasi correlation, and the Bowring correlation. These nine model configurations were trained in two data scenarios: one with plentiful training points (7,350), and the other with a severely restricted training set (9 points). Using this experimental structure, an ML-base model combination may be identified as having the most favorable performance, uncertainty behavior, and resistance to performance degradation in scenarios of limited data.

\subsection{Data Preparation} \label{sec:methods_data}

The CHF dataset, filtered for instances of DO within the ranges of validity for the Biasi and Bowring correlations, was first shuffled to homogenize the entries and ensure adequate coverage of the data space. The data in the input channels and the outputs were then standardized (i.e., via Z-score normalization) so that their distributions would have a mean of 0 and a standard deviation of 1 \cite{garcia2015data}. This standardization is necessary to avoid biasing the model during training, which may attribute higher importance to input channels of larger magnitude. The resulting data array was partitioned into training, validation, and testing partitions in an 80\%/10\%/10\% split. The validation set is used for the optimization of network structure and other hyperparameters, and is separate from the testing partition, which is used to evaluate the final trained model. The use of testing data in the process of hyperparameter optimization may inadvertently transfer some knowledge of the holdout data to the model during tuning and thereby cause inflated performance metrics \cite{boulesteix2015ten}. These data partitions and their contents were identical for every model used in this study to ensure an appropriate comparison. 

A second dataset was constructed by simply reducing the training partition's size to only nine points while maintaining the size and contents of the validation and testing partitions. The same model configurations used in the plentiful data scenario would be trained on this limited data scenario. The purpose of this scenario is to investigate the hybrid approach's resistance to performance degradation due to poor training data compared to that of a pure ML model. The training set's size in this limited case represents \textbf{0.1\%} of the complete filtered dataset, compared with the previous scenario's training set consisting of \textbf{80\%}.
Future development of hybrid models with limited data, for actual use, would require more comprehensive qualification.

\subsection{Deep Neural Network Ensembles} \label{sec:methods_ensemble}

The concept of ensembling is based on the idea that a committee of unique DNN models will outperform a single model \cite{lakshminarayanan2017simple}. Because DNNs are purely deterministic once trained, they will produce the same value every time a prediction is made on the same input vector. When multiple unique models are well trained with different hyperparameters and optimizations, they will each produce slightly different predictions with the same input, allowing for their treatment as samples from an underlying distribution \cite{yaseen2023quantification}. Measures of these distributions, such as the means and standard deviations for every point in the test dataset, can then be computed. The confidence and uncertainty in the model parameters can then be quantitatively inferred from this analysis \cite{tan2023single}.

The simplest implementation of a DNN ensemble is using an initialization-based strategy, in which a set of models is trained using different weight/bias starting values \cite{lecun2015deep}. Each of these models subsequently navigates a different solution space to attempt to find a global minimum in the objective function's space. In this study, a set of 20 otherwise identical models were trained with different random number generator seeds using Google's TensorFlow \cite{tensorflow2015-whitepaper}. The network architecture consisted of seven densely connected hidden layers, using the Adam optimizer and a mean-square-error objective function. An exponential learning rate decay with a 0.96 constant was also implemented to more finely navigate the objective function's loss space in later training passes (epochs). The network depth, as well as hyperparameters such as the number of neurons per layer and the activation functions, were optimized using the validation partition with 1000 randomized prospective configurations. RayTune\cite{liaw2018tune} was used for this process, along with the asynchronous successive halving algorithm, to reduce computational expenses. Once the final architecture had been configured, the ensemble models were trained to 250 epochs and were then used to predict the outputs of the test dataset.

For each input combination in the test dataset, the 20 ensemble models produced 20 unique predictions. With these 20 samples, the means and standard deviations were collected. These mean values were used as the ensemble's ``single-value'' predictions in further analysis, with the standard deviations representing their uncertainties.

\subsection{Bayesian Neural Networks} \label{sec:methods_bnn}

The second method, BNNs, extends the framework of DNNs but with built-in uncertainty estimation. Instead of using single-valued weights, BNNs use probability distributions. This approach allows the model to directly capture uncertainty in the tunable parameters \cite{mackay1992practical}. During training, BNNs use techniques such as variational inference to approximate the posterior distribution over the weights, providing a structured way to quantify uncertainty. Variational inference transforms the problem of exact Bayesian inference into an optimization problem, which enables the network to learn a distribution of weights that best fits the data. This allows BNNs to express uncertainty about predictions (especially in regions with limited data) and be more generalizable by considering multiple plausible models rather than a single set of weights.

The BNN was implemented using TensorFlow Probability and was constructed from four densely connected hidden layers using the Flipout estimator. These layers introduce epistemic uncertainty because they estimate a distribution over the weights and biases instead of considering them to be single values. The Flipout estimator reduces the variance in gradient estimates during training by performing a Monte Carlo approximation. The network architecture outputs two parameters defining a normal distribution: the mean and scale (standard deviation) of the target values. During training, the negative log likelihood of this distribution is minimized, which also incorporates the Kullback--Leibler divergence between prior and posterior distributions. For hyperparameter optimization, the procedure from Section \ref{sec:methods_ensemble} was implemented.

The prior and posterior distributions for the weights were modeled as normal distributions, the prior having a mean of 0 and variance of 1, and the posterior initialized with small perturbations from the prior. Variational inference was used to optimize the posterior distribution by approximating the true distribution over the weights. The model was trained to 500 epochs using the Adam optimizer with the same learning rate decay scheme as from Section \ref{sec:methods_ensemble}. In a BNN, predictions are made from a trained model by sampling from the posterior distribution, producing unique values even when predicting on an identical input vector. For this study, 200 samples were taken from the posterior predictive distribution for every input entry in the test dataset; as with the previously discussed ensemble method, the means and standard deviations of these distributions were interpreted as the final predictions and uncertainties. The specific number of samples needed per input set was determined via a convergence study to ensure the stability of the reported values.

\subsection{Deep Gaussian Processes} \label{sec:methods_dgp}

Although DGPs also use a hierarchical network structure, they are layers of Gaussian processes (GPs). GPs are a finite collection of random variables that have a joint multivariate normal distribution. For any given input, the mean and variance are computed for a target distribution, where the mean is considered to be the prediction. GPs use kernel functions to compute the covariance between data points, making them especially useful in cases of limited data. Structuring GPs with ``depth'' in layers extends their ability to capture more complex representations of a dataset. Coupling this hierarchical structure with GPs allows them to effectively extract uncertainties from complex data \cite{damianou2013deep}. DGPs may also be configured to capture aleatoric uncertainty via the observation noise variance within each layer, if desired. TensorFlow Probability was again used to implement the DGP models. The network architecture consisted of two layers of variational Gaussian processes using the Radial Basis Function kernel to define the covariance structure. This model was then trained to 500 epochs.

\section{Results} \label{sec:results}

For each of the three ML methods used in this study, a purely data-driven case was run without a base model, and two hybrid cases were then run with the Biasi and Bowring base models. This yielded a total of nine unique model configurations, which were trained and tested in two data scenarios: plentiful training data and limited training data. 

Quantifying and comparing the performance between the three modeling approaches (pure ML, Biasi hybrid, and Bowring hybrid) in each case is primarily accomplished using a set of seven metrics. The terms $\upmu_{\mathrm{error}}$ and $\mathrm{Max}_{\mathrm{error}}$ correspond to the mean and maximum of the absolute relative error distribution across the final predictions. The relative standard deviations (rStd) describe the shape of the distributions along each input predictions' samples to be a measure of uncertainty for each of the final predictions. In the case of the DGP models, the standard deviations are directly extracted from the GP layers. Another error metric, relative root-mean-square error (rRMSE), is included as a more sensitive measure that penalizes larger error values to a higher degree and is scale-invariant. The definition of rRMSE has been inconsistent over several literature entries, but this study uses the definition in the OECD/NEA benchmark report, as represented in Eq.~(\ref{eqn:rrmse}) \cite{lecorre2023benchmark}:

\begin{equation} \label{eqn:rrmse}
    \text{rRMSE}\; (\%) = \sqrt{\frac{1}{N}\sum_{i = 1}^N \left(\frac{\hat{y}_i - y_i}{y_i}\right)^2} \times 100 \%
\end{equation}

The metric $F_{\text{error}}>10\%$ corresponds to the percentage of absolute relative error values that are larger than 10\%. Finally, as the last metric, $R^2$ is the coefficient of determination, which ranges from 0 to 1; an ideal model would attain a value of 1.

\subsection{DNN Ensembles} \label{sec:results_ensemble}

After all 20 of the ensemble models were run for each of the experimental configurations, the seven metrics were computed. These metrics are listed in Table \ref{tab:comparison_ensemble}, and each row's most favorable value is highlighted in gray. For the 80\% training data case, all three approaches obtain $\upmu_{\mathrm{error}}$ values below 2\%, and the hybrid models present slightly smaller values in comparison with those of the pure ML approach. The maximum error values, however, are larger for the hybrid models in comparison with the pure ML model's $36.03$\%. This difference is mild, and in assessing the fraction of these errors above 10\%, the hybrid models are observed to have the smallest values (0.871\% of the pure ML model in comparison with 0.654\% and 0.764\% of Biasi and Bowring, respectively). All three approaches obtain a high $R^2$ above 0.99, indicating strong agreement between predicted and measured values. In terms of uncertainty, all three ensembles exhibit a high degree of confidence (all below 1.55\%), with the Bowring hybrid variant having the smallest relative standard (1.277\%).

In the case of the 0.1\% training size (using only nine points for training), the hybrid ensembles have significantly more favorable metrics in comparison with the pure ML ensemble, especially for the Biasi hybrid variant. The $\upmu_{\mathrm{error}}$ values for all three ensembles are larger than those for the 80\% training size case, with the pure ML value changing from 1.947\% to a far larger 35.42\%. The hybrids' $\upmu_{\mathrm{error}}$ values increase by just under 4.5 percentage points but remain below those of the stand-alone correlations (6.935\% for Biasi and 6.778\% for Bowring). Compared with the stand-alone correlations, the hybrid ensembles attain more favorable values in every metric, even with only nine training points. In the case of the pure ML ensemble, the limited training data significantly impede learning and the ability to generalize to new data.

\renewcommand{\arraystretch}{0.9}
\begin{table}[ht!]
    \centering
    \caption{Comparison of the ensemble performance metrics between the pure and hybrid models}
    \label{tab:comparison_ensemble}
    \begin{tabular}{lccc}
        \toprule
        Metric & Pure ML & Biasi & Bowring \\
         & DNN & Hybrid & Hybrid \\
        \midrule
        \multicolumn{4}{c}{\textbf{$\mathbf{7{,}350}$ Training Points ($\mathbf{80}$\% Case)}} \\
        \midrule
        $\upmu_\text{error}$ & $1.947\%$ & \cellcolor{gray!25}$1.846\%$ & $1.883\%$ \\ \hline
        $\text{Max}_{\text{error}}$ & \cellcolor{gray!25}$36.03\%$ & $37.54\%$ & $41.25\%$ \\ \hline
        $\text{Mean rStd}$ & $1.541\%$ & $1.302\%$ & \cellcolor{gray!25}$1.277\%$ \\ \hline
        $\text{Max rStd}$ & $8.967\%$ & \cellcolor{gray!25}$7.231\%$ & $11.89\%$ \\ \hline
        \text{rRMSE} & $3.023\%$ & \cellcolor{gray!25}$2.793\%$ & $2.940\%$ \\ \hline
        $F_{\text{error}}>10\%$ & $0.871\%$ & \cellcolor{gray!25}$0.654\%$ & $0.764\%$ \\ \hline
        $R^2$ & \cellcolor{gray!25}$0.9970$ & $0.9966$ & $0.9963$  \\
        \midrule
        \multicolumn{4}{c}{\textbf{$\mathbf{9}$ Training Points ($\mathbf{0.1}$\% Case)}} \\
        \midrule
        $\upmu_\text{error}$ & $35.42\%$ & \cellcolor{gray!25}$6.082\%$ & $6.354\%$ \\ \hline
        $\text{Max}_{\text{error}}$ & $323.9\%$ & $59.10\%$ & \cellcolor{gray!25}$57.82\%$ \\ \hline
        $\text{Mean rStd}$ & $23.52\%$ & \cellcolor{gray!25}$1.930\%$ & $2.670\%$ \\ \hline
        $\text{Max rStd}$ & $53.09\%$ & \cellcolor{gray!25}$24.62\%$ & $25.43\%$ \\ \hline
        \text{rRMSE} & $59.10\%$ & \cellcolor{gray!25}$8.804\%$ & $9.541\%$ \\ \hline
        $F_{\text{error}}>10\%$ & $70.81\%$ & \cellcolor{gray!25}$17.10\%$ & $18.85\%$ \\ \hline
        $R^2$ & $0.5405$ & \cellcolor{gray!25}$0.9782$ & $0.9739$  \\
        \bottomrule
    \end{tabular}
\end{table}

\begin{figure}[ht!]
    \centering
    \includegraphics[width=\linewidth]{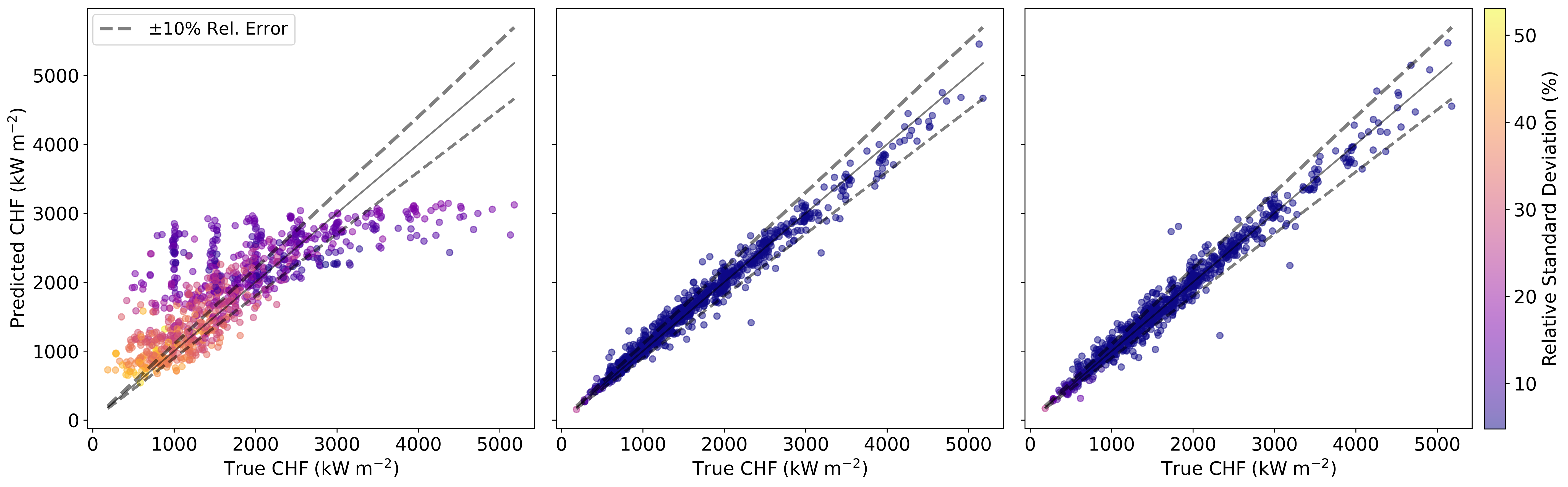}
    \begin{subfigure}[b]{0.32\textwidth}
        \caption{Pure DNN}
    \end{subfigure}
    \begin{subfigure}[b]{0.32\textwidth}
        \caption{Biasi hybrid}
    \end{subfigure}
    \begin{subfigure}[b]{0.32\textwidth}
        \caption{Bowring hybrid}
    \end{subfigure}
    \caption{Parity plots of the pure ML and hybrid ensembles with uncertainty overlay for the 0.1\% training size case.}
    \label{fig:parity_ensemble_001p}
\end{figure}

To graphically show how the three ensembles perform in the limited training data scenario, their parity plots are included in Figure \ref{fig:parity_ensemble_001p}. These plots compare the predicted CHF value with the actual CHF value to display how ``wrong'' each prediction is compared with a perfect prediction along the identity line. These values are from the test dataset, which is still 10\% of the original filtered dataset. The relative standard deviation on a common scale is also overlaid to show any potential trends in uncertainty.

The first behavior of note is for the pure ML ensemble, which exhibits significant deviation from the ideal line, especially in attempting to predict larger CHF values. The larger \textit{relative} errors, however, are located in attempting to predict the smaller CHF values, as shown by the $\pm 10\%$ error bars. This larger region of relative errors also corresponds to a higher relative standard deviation, which is more favorable behavior than the converse. The most expressive comparison is between the pure ML ensemble's parity plot and those of the hybrid ensembles. Both the Biasi and Bowring hybrids show tight adherence to the identity line; a large majority of the points are contained within the relative error bounds (only 17.10\% and 18.85\% lie outside for the Biasi and Bowring hybrids).

\begin{figure}[ht!]
    \centering
    \begin{subfigure}{0.49\textwidth}
        \centering
        \includegraphics[width=\linewidth]{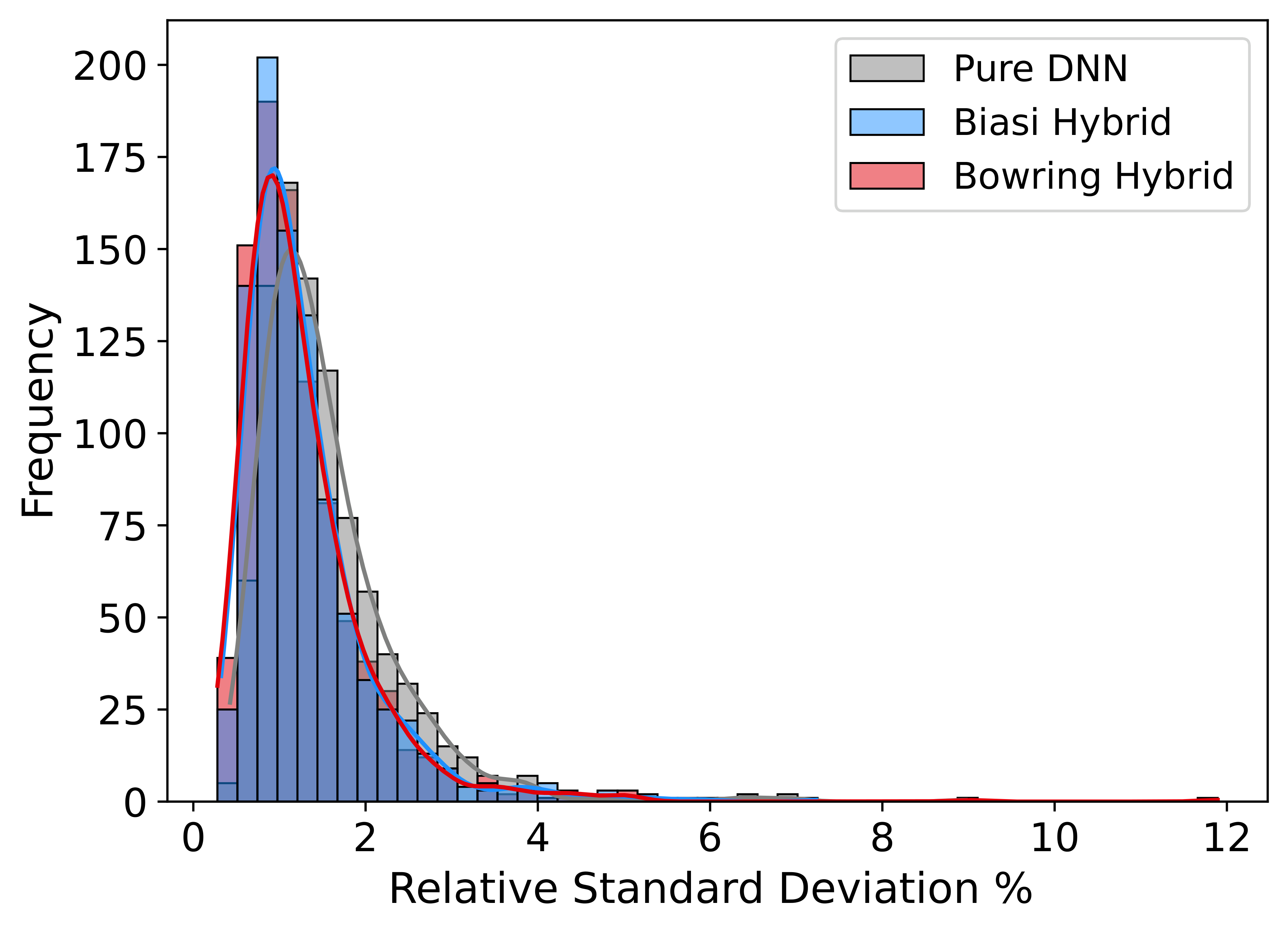}
        \caption{80\% Training Size}
    \end{subfigure}
    \begin{subfigure}{0.49\textwidth}
        \centering
        \includegraphics[width=\linewidth]{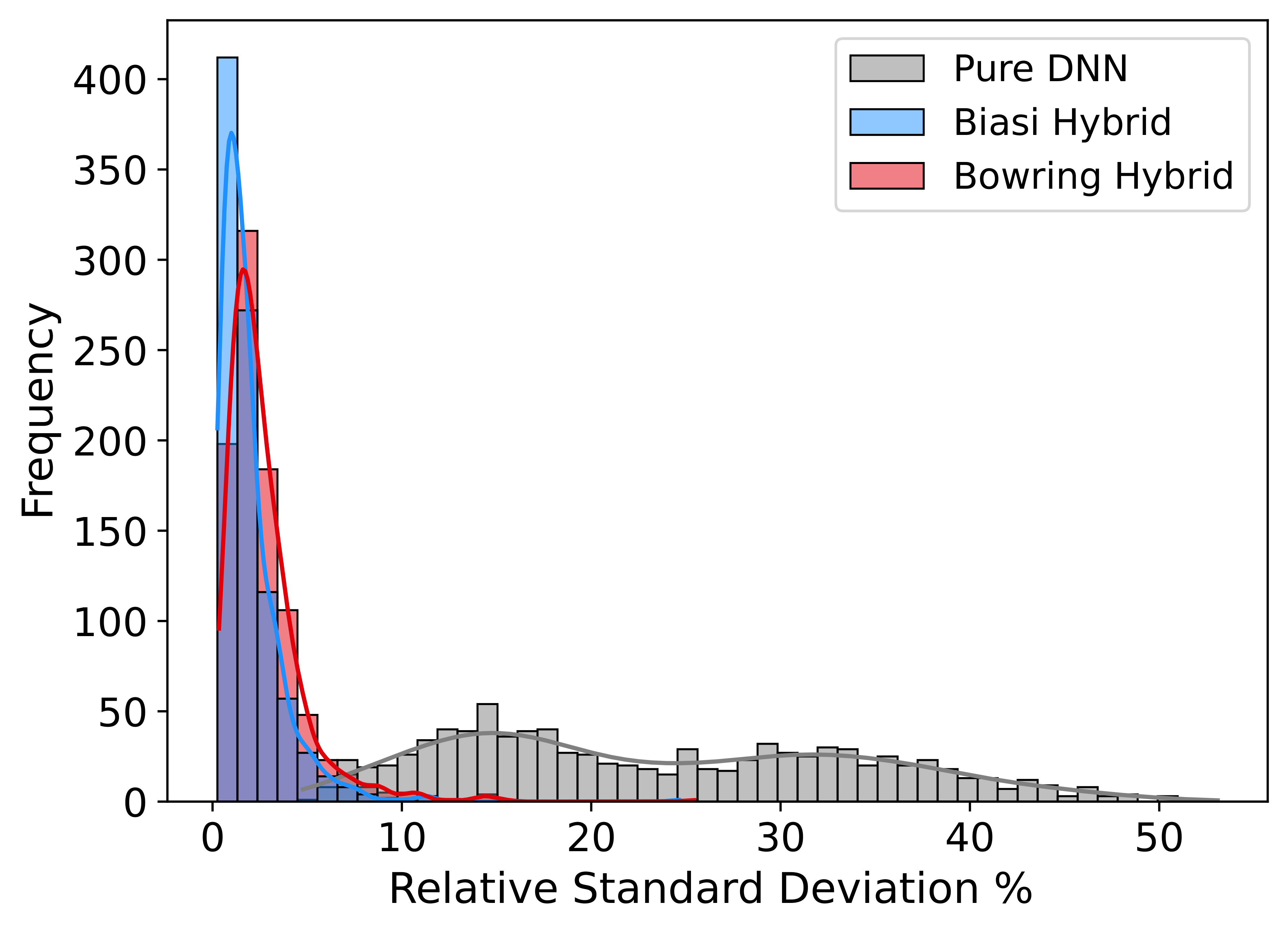}
        \caption{0.1\% Training Size}
    \end{subfigure}
    \caption{Distributions of the rStds for the ensemble approach: histograms and KDE curves.}
    \label{fig:rstd_dist_ensemble}
\end{figure}

The distributions of the relative standard deviations were then plotted for the three ensemble variants in both the plentiful training data scenario and the limited scenario, as shown in Figure \ref{fig:rstd_dist_ensemble}. These plots were constructed with equal-width histograms and kernel density estimation (KDE). In the plentiful data scenario, all three ensembles perform similarly, albeit with the pure ML ensemble having a slightly higher mean and spread. Although there are larger rStd values between 7\% and 12\%, these are located far away from the bulk of values. In the 0.1\% training size scenario, the Biasi and Bowring hybrid ensembles maintain their overall shapes with some increase in mean values, but with both mean rStd values remaining under 2.7\%. The pure ML ensemble's distribution shows a significant rise in overall rStd values; the distribution takes a bimodal shape and a maximum of 53.09\%.

\begin{figure}[ht!]
    \centering
    \begin{subfigure}{0.49\textwidth}
        \centering
        \includegraphics[width=\linewidth]{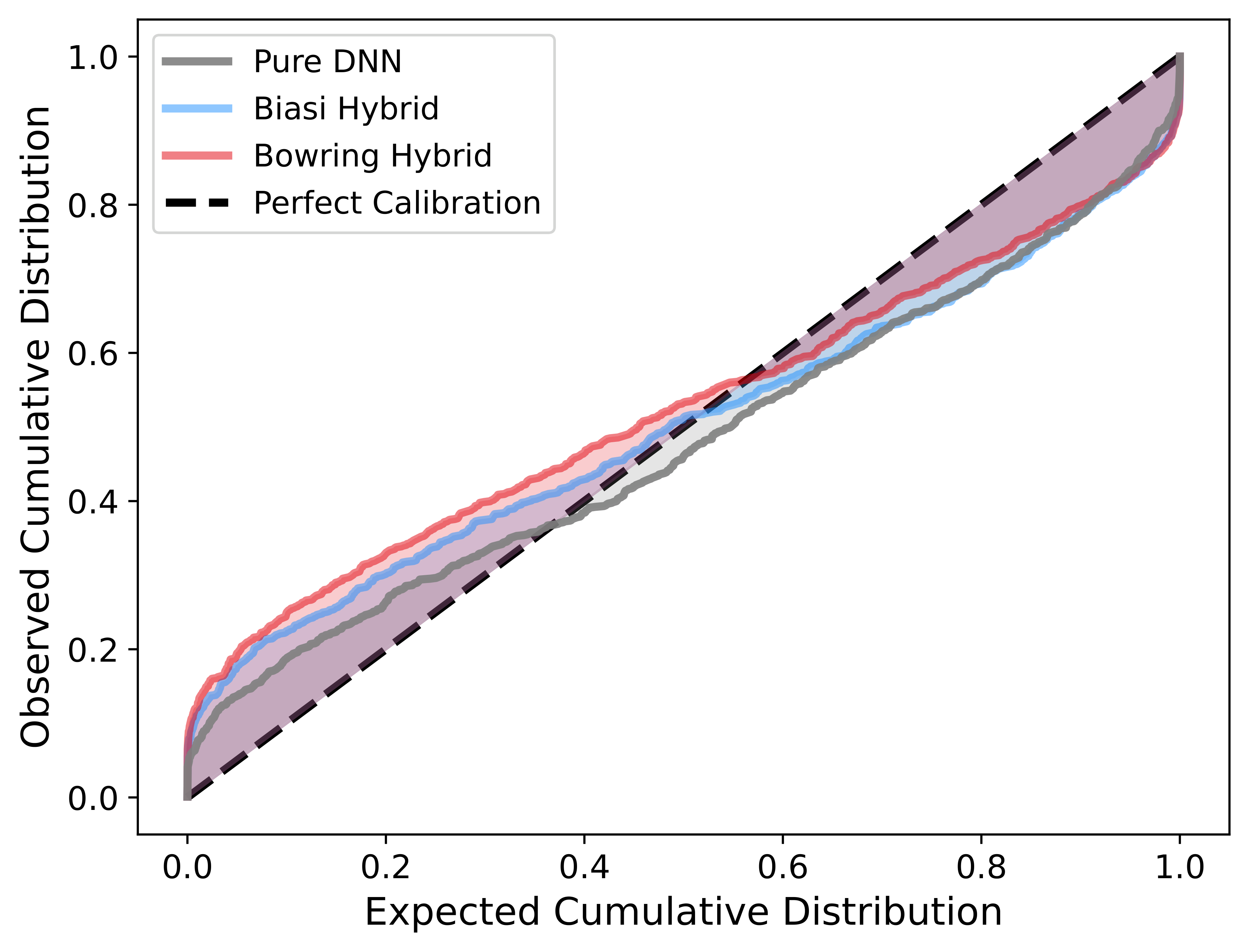}
        \caption{80\% Training Size}
    \end{subfigure}
    \begin{subfigure}{0.49\textwidth}
        \centering
        \includegraphics[width=\linewidth]{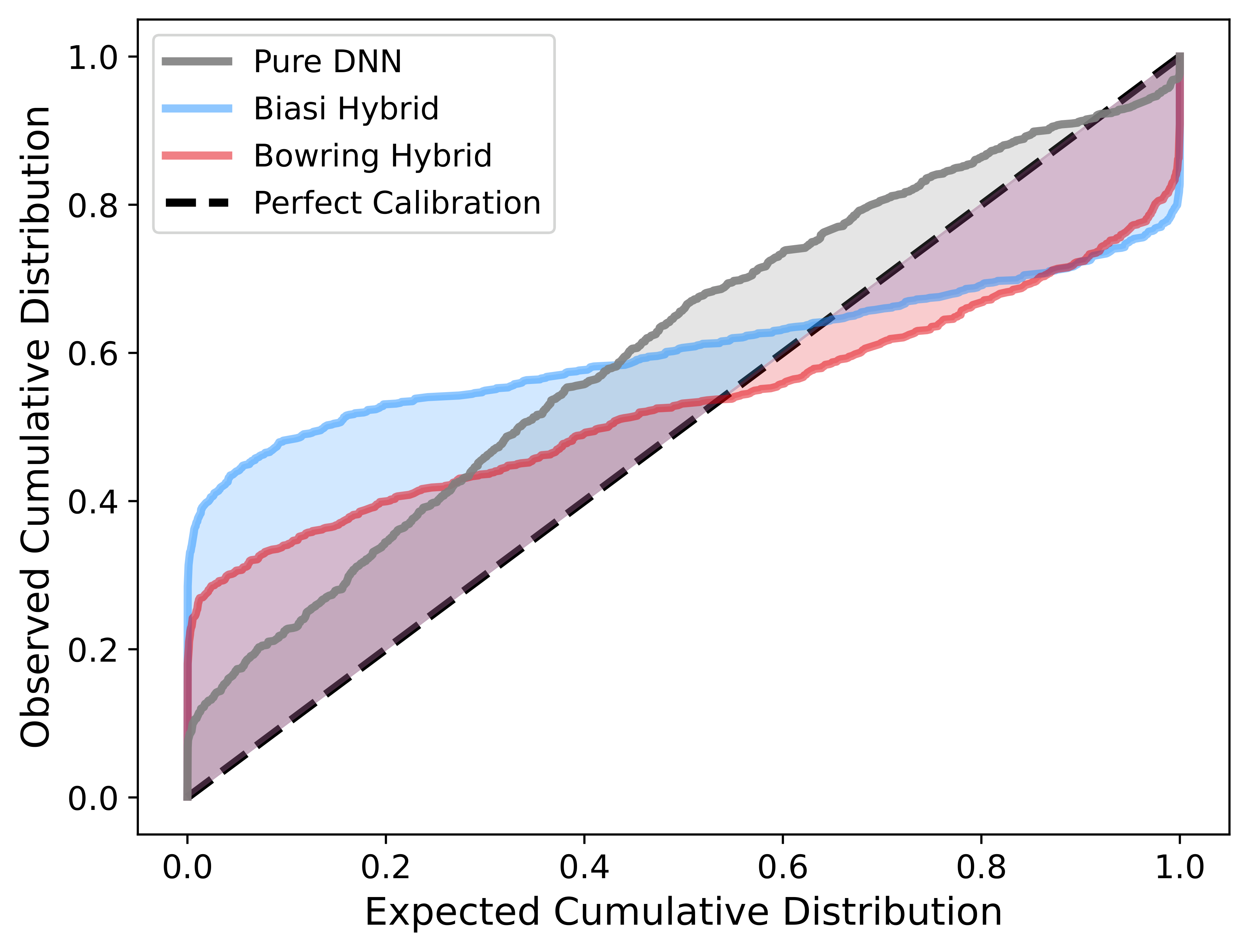}
        \caption{0.1\% Training Size}
    \end{subfigure}
    \caption{Calibration curves for the DNN ensemble.}
    \label{fig:calibration_ensemble}
\end{figure}

Evaluating the calibration of each model is also necessary. This was achieved using calibration curves and miscalibration areas. The miscalibration area is simply the area between a given curve and the identity line, which provides a quantitative aspect to this otherwise graphical analysis. Each of the three ensemble variants is plotted for both training size scenarios in Figure \ref{fig:calibration_ensemble}. In the 80\% case, all three of the curves adhering to the ideal line indicates that these ensembles' predictions are well calibrated; this is further supported by relatively small miscalibration areas (0.0679, 0.0771, and 0.0824 for the pure ML, Biasi, and Bowring ensembles, respectively). In the limited data scenario, all three of the ensembles' predictions exhibit poorer calibrations, especially in the case of Biasi. The pure ML ensemble maintains the most favorable calibration of the three, although it is entirely above the perfect calibration line. As discussed in Section \ref{subsec:uq_quality}, this indicates that uncertainty estimates are disproportionately large in comparison with the expected distribution, indicating underconfidence. Both of the hybrid models show the opposite effect; their estimates are smaller than expected, indicating overconfidence. Each of the miscalibration areas is larger than those of the 0.1\% case: 0.113, 0.188, and 0.135 for pure ML, Biasi, and Bowring ensembles, respectively.

Both the Biasi and Bowring hybrid ensemble variants have demonstrated notable performance in accuracy and in their uncertainty estimates in comparison with those of the pure ML ensemble. Most significantly, these hybrid ensembles show a high degree of resistance to limited training datasets, achieving more favorable error metrics than even the stand-alone empirical correlations that they are based on. Although a guarantee that a hybrid approach will always perform better than or equal to that of the base model is impossible, the above results support that, at least in this case, this behavior is true.

\subsection{Bayesian Neural Networks}

The second set of cases uses BNNs to quantify the prediction uncertainties, again for the pure ML and two hybrid variants in both data scenarios. For each of the input vectors, 200 ``samples'' were predicted for the uncertainty estimation. The seven performance metrics were computed. These metrics are presented in Table \ref{tab:comparison_BNN}. The trends noted for the DNN ensemble in Section \ref{sec:results_ensemble} are present here; all three models of the 80\% training size scenario behave similarly. All these values, compared with the ensemble's, are larger in error with a larger degree of uncertainty. In the limited data scenario, there is the expected jump in error in the pure ML model's predictions but with a larger degree than that of the ensemble's (2.818\% up to 89.58\% $\upmu_{\mathrm{error}}$). This is also the case for the six other metrics (especially with a negative $R^2$), signaling a complete loss of generalization to new data. Both the Biasi and Bowring hybrid models show an expected rise in error metrics, but the Biasi hybrid exhibits values that are larger than those of the stand-alone Biasi correlation. This behavior is entirely different from the Biasi hybrid ensemble's behavior, which approached but did not exceed the error of the stand-alone correlation. The Bowring hybrid BNN model, however, remains within the performance metrics of the stand-alone correlation. Despite this decreased performance of the Biasi hybrid BNN, it reports the highest confidence in its predictions, indicating that it is not properly calibrated.

\begin{table}[ht!]
    \centering
    \caption{Comparison of the BNN performance metrics between the pure and hybrid models}
    \label{tab:comparison_BNN}
    \begin{tabular}{lccc}
        \toprule
        Metric & Pure ML & Biasi & Bowring \\
         & BNN & Hybrid & Hybrid \\
        \midrule
        \multicolumn{4}{c}{\textbf{$\mathbf{7{,}350}$ Training Points ($\mathbf{80}$\% Case)}} \\
        \midrule
        $\upmu_\text{error}$ & $2.818\%$ & \cellcolor{gray!25}$2.803\%$ & $2.909\%$ \\ \hline
        $\text{Max}_{\text{error}}$ & $34.28\%$ & \cellcolor{gray!25}$29.06\%$ & $38.05\%$ \\ \hline
        $\text{Mean rStd}$ & $3.848\%$ & $3.897\%$ & \cellcolor{gray!25}$3.833\%$ \\ \hline
        $\text{Max rStd}$ & $24.60\%$ & $17.27\%$ & \cellcolor{gray!25}$15.83\%$ \\ \hline
        \text{rRMSE} & \cellcolor{gray!25}$4.109\%$ & $4.118\%$ & $4.311\%$ \\ \hline
        $F_{\text{error}}>10\%$ & $2.503\%$ & \cellcolor{gray!25}$2.067\%$ & $2.612\%$ \\ \hline
        $R^2$ & $0.9926$ & \cellcolor{gray!25}$0.9929$ & $0.9914$ \\
        \midrule
        \multicolumn{4}{c}{\textbf{$\mathbf{9}$ Training Points ($\mathbf{0.1}$\% Case)}} \\
        \midrule
        $\upmu_\text{error}$ & $89.58\%$ & $9.573\%$ & \cellcolor{gray!25}$6.474\%$ \\ \hline
        $\text{Max}_{\text{error}}$ & $1069.\%$ & $72.91\%$ & \cellcolor{gray!25}$60.29\%$ \\ \hline
        $\text{Mean rStd}$ & $57.79\%$ & \cellcolor{gray!25}$12.08\%$ & $15.49\%$ \\ \hline
        $\text{Max rStd}$ & $72.87\%$ & \cellcolor{gray!25}$66.54\%$ & $110.5\%$ \\ \hline
        \text{rRMSE} & $138.0\%$ & $13.13\%$ & \cellcolor{gray!25}$9.616\%$ \\ \hline
        $F_{\text{error}}>10\%$ & $88.36\%$ & $35.80\%$ & \cellcolor{gray!25}$18.50\%$ \\ \hline
        $R^2$ & $-0.4934$ & $0.9674$ & \cellcolor{gray!25}$0.9701$ \\
        \bottomrule
    \end{tabular}
\end{table}

\begin{figure}[ht!]
    \centering
    \includegraphics[width=\linewidth]{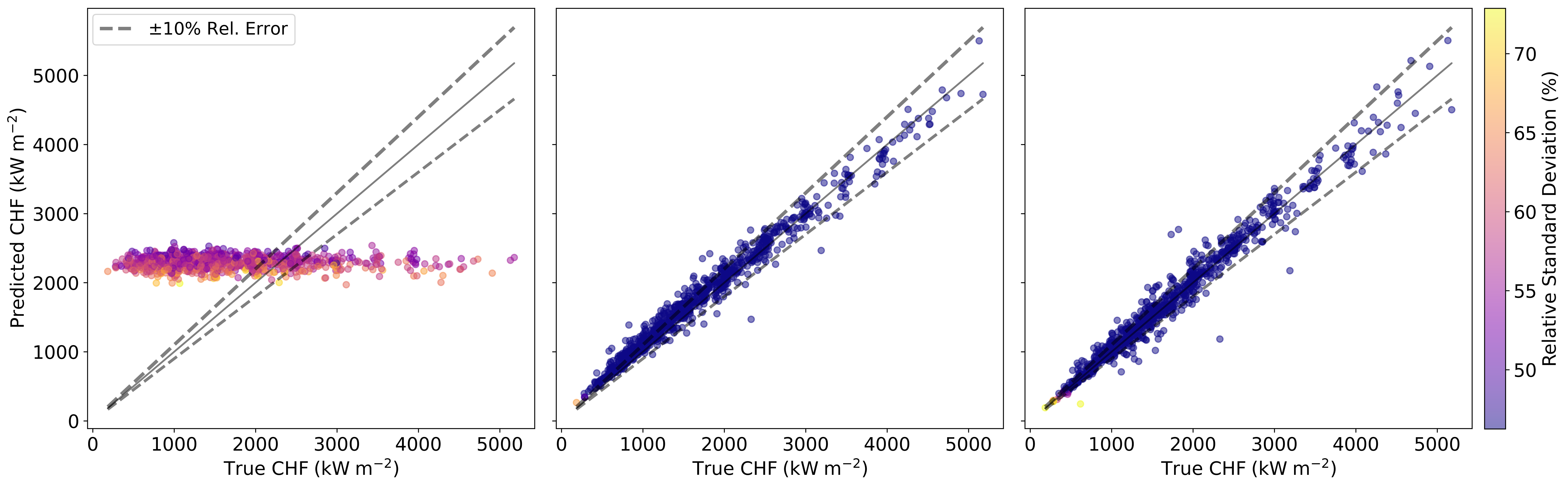}
    \begin{subfigure}[b]{0.32\textwidth}
        \caption{Pure DNN}
    \end{subfigure}
    \begin{subfigure}[b]{0.32\textwidth}
        \caption{Biasi}
    \end{subfigure}
    \begin{subfigure}[b]{0.32\textwidth}
        \caption{Bowring}
    \end{subfigure}
    \caption{Parity plots of the pure ML and hybrid BNN with uncertainty overlay for the 0.1\% training size~case.}
    \label{fig:parity_bnn_001p}
\end{figure}

Figure \ref{fig:parity_bnn_001p} presents the parity plots for these BNN models in the limited data scenario. The collapse of the pure ML model's performance is shown. The model predicts all values about a near-horizontal axis around 2,250 \si{\kilo\watt\per\square\meter}. This indicates that the model fails entirely to generalize to the hidden testing dataset, effectively becoming useless in its application. The uncertainty estimates in this case are larger for smaller predicted CHF values, with a negative gradient with increasing predicted values. The two hybrid models are shown with the correct trends and relatively tight grouping around the identity line; some positive bias exists in the Biasi hybrid's predictions at lower true CHF values. 

\begin{figure}[ht!]
    \centering
    \begin{subfigure}{0.49\textwidth}
        \centering
        \includegraphics[width=\linewidth]{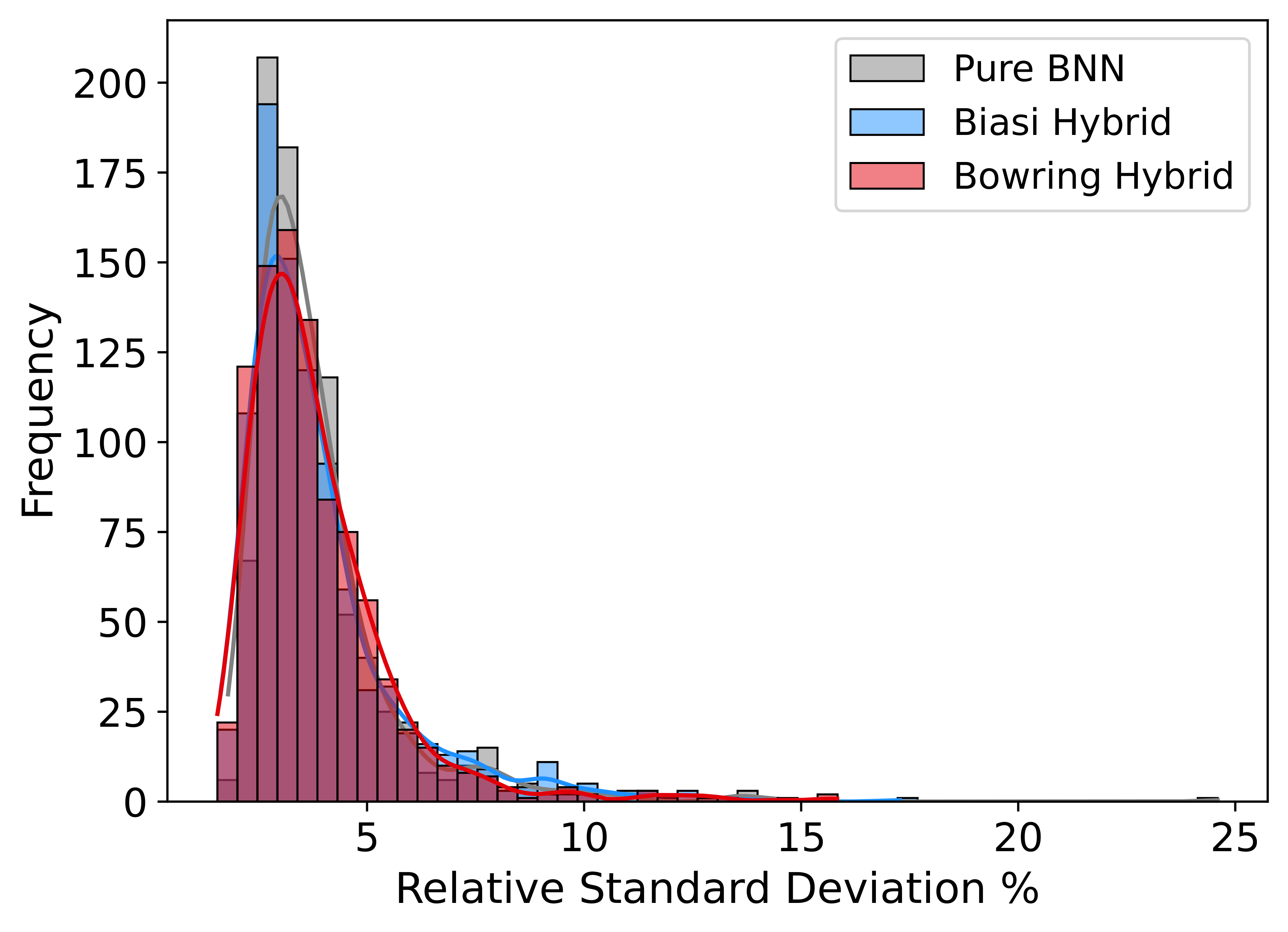}
        \caption{80\% Training Size}
    \end{subfigure}
    \begin{subfigure}{0.49\textwidth}
        \centering
        \includegraphics[width=\linewidth]{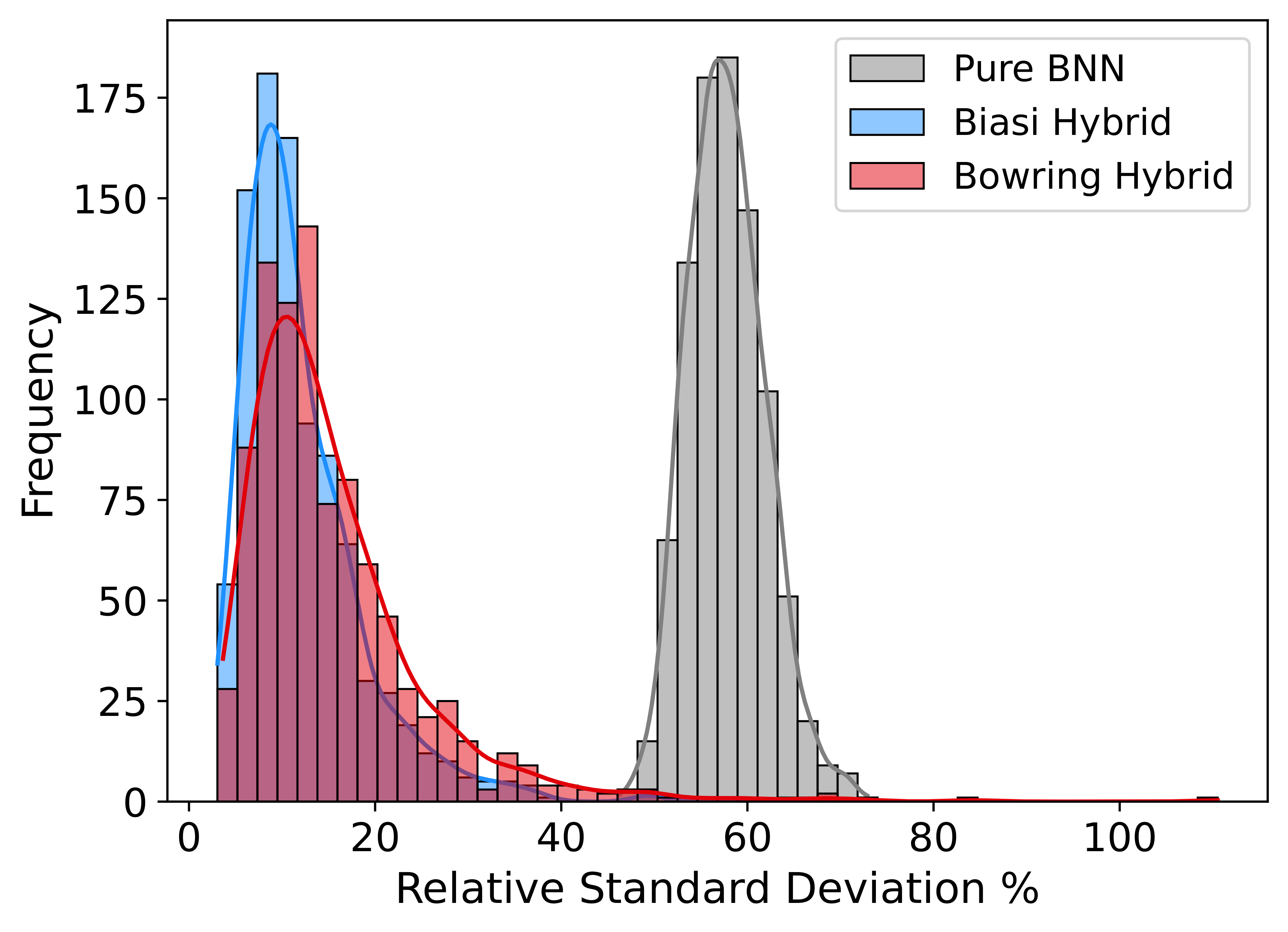}
        \caption{0.1\% Training Size}
    \end{subfigure}
    \caption{Distributions of the rStds for the BNN approach: histograms and KDE curves.}
    \label{fig:rstd_dist_bnn}
\end{figure}

To show a more focused look at the uncertainty estimates from each of these models, the relative standard deviations are provided in Figure \ref{fig:rstd_dist_bnn}. As observed in the performance metrics, the means and shapes of the models' distributions in the plentiful training data scenario are similar. There are no significant deviations, except for the outlier just below 25\%, which belongs to the pure ML model. In the limited data scenario, there is a clear difference between the distributions of the hybrid models and that of the pure ML model. The pure ML model's rStd values are near symmetric at about the 57.79\% mean value and are a distinct peak in comparison with the hybrid model distributions in the left-hand corner. This behavior is favorable to some degree because the very large error associated with this model is paired with nonconfidence in its predictions.

\begin{figure}[ht!]
    \centering
    \begin{subfigure}{0.49\textwidth}
        \centering
        \includegraphics[width=\linewidth]{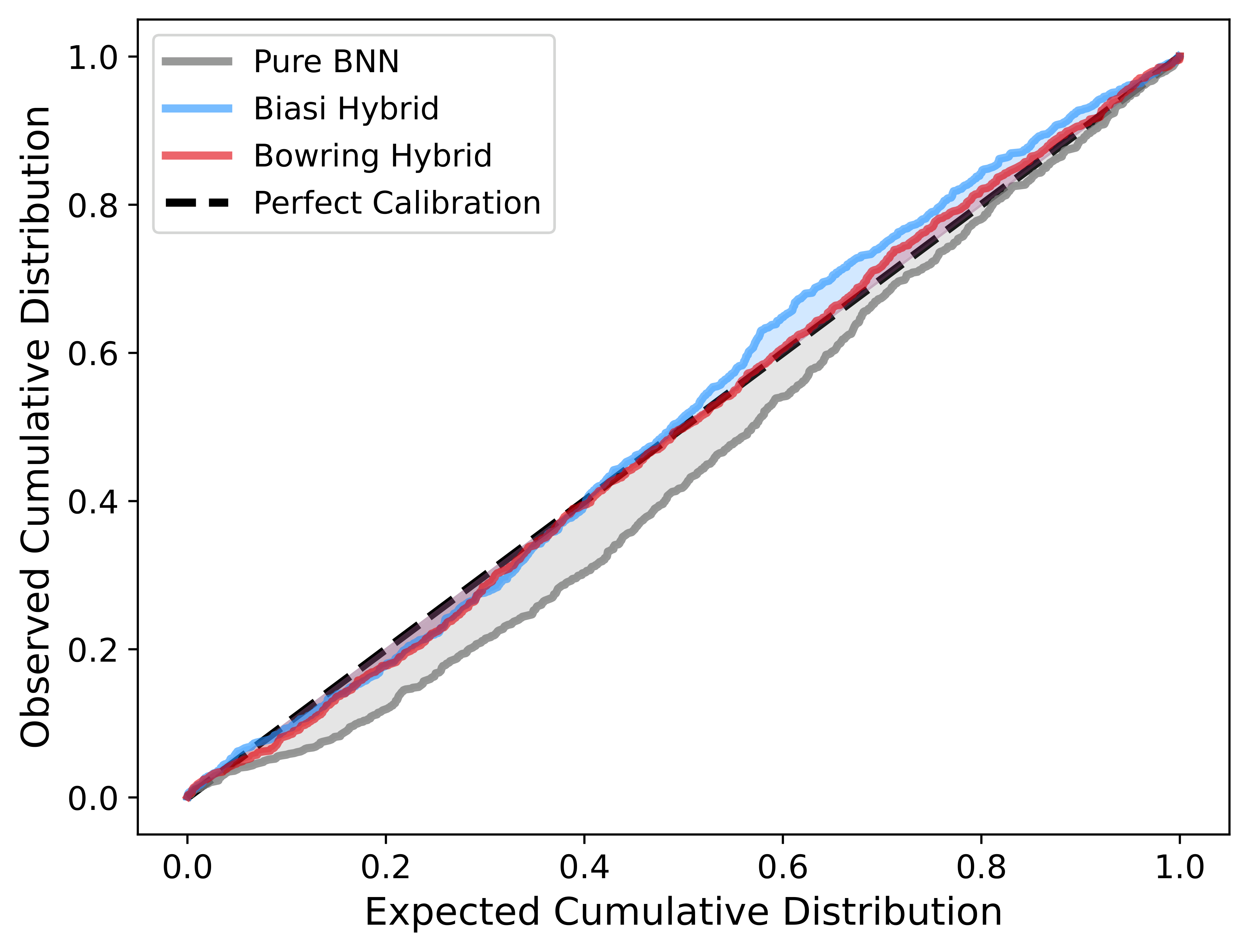}
        \caption{80\% Training Size}
    \end{subfigure}
    \begin{subfigure}{0.49\textwidth}
        \centering
        \includegraphics[width=\linewidth]{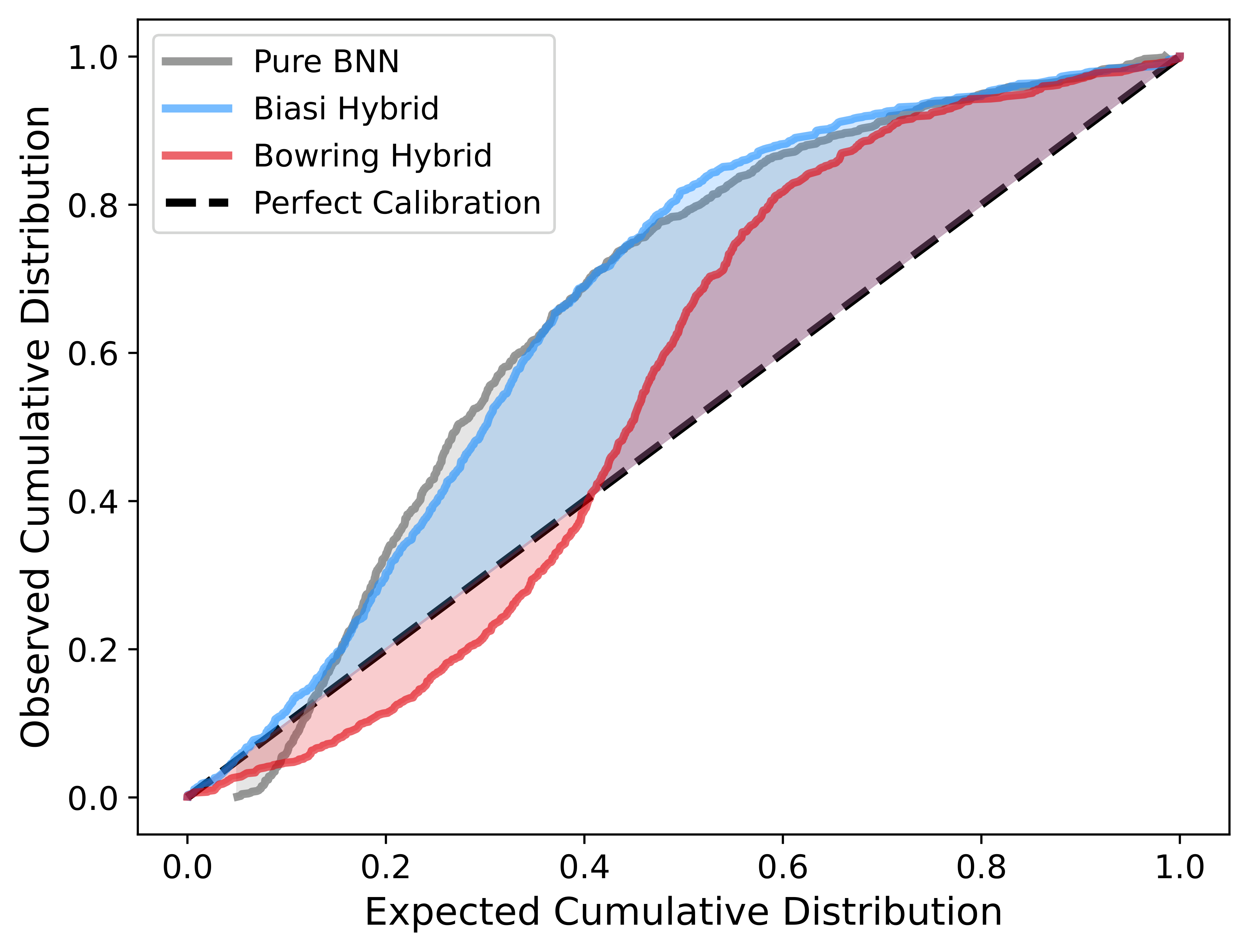}
        \caption{0.1\% Training Size}
    \end{subfigure}
    \caption{Calibration curves for the BNN approach.}
    \label{fig:calibration_bnn}
\end{figure}

Figure \ref{fig:calibration_bnn} shows the calibration curves for the BNN approach. Compared with the DNN ensemble predictions in the plentiful training scenario (discussed in Section \ref{sec:results_ensemble}), the BNN predictions in the plentiful training scenario are extremely well calibrated; the hybrid models visibly align with the identity line. The pure ML model also indicates adequate calibration, having a miscalibration area of 0.0506 (compared with the ensemble counterpart's 0.0679). Both hybrid models' curves also feature smaller miscalibration areas: 0.0233 and 0.0115 for the Biasi and Bowring variants, respectively. This behavior is starkly contrasted by that of the limited data scenario, with miscalibration areas of 0.1691, 0.1669, and 0.1023 for the pure ML, the Biasi hybrid, and the Bowring hybrid, respectively. All three of the model variants systematically overestimate uncertainties, leading to large prediction intervals (underconfidence). The Bowring hybrid has the most favorable calibration of the three, albeit with significant underconfidence.

\subsection{Deep Gaussian Processes}

Each of the DGP models was trained and tested using the same test dataset used for the other models. The standard deviations were extracted directly from the model instead of using a sampling approach such as the approaches used for the DNN ensembles and the BNN models. The same performance metrics were then computed. These metrics are summarized in Table \ref{tab:comparison_dgp}. In comparison with the performance of the previous two methods, nearly every error metric is significantly larger. Between the DGP models themselves, the Biasi hybrid model outperforms both the pure ML and Bowring hybrid models by a significant margin. Although it does not match the performance of the Biasi hybrid, the Bowring hybrid model also performs with more favorable metrics in comparison with the pure ML model. The exception in this case is in the Bowring hybrid's larger relative standard deviation values, which indicate that more uncertainty is associated with those predictions. 

In the limited training dataset case (using only nine points for training), all three of the models' predictions were observed with expected deterioration in all metrics. The two hybrid models have notably smaller error metrics compared with those of the pure ML model, except for the maximum relative standard deviation. Of the three models, the Bowring hybrid produced the most accurate predictions with the smallest uncertainties on average. Notably, as with the pure ML BNN models, the pure ML DGP predicts nearly the same value for every input combination, indicating a complete inability to generalize to new data.

\begin{table}[ht!]
    \centering
    \caption{Comparison of the DGP performance metrics between the pure and hybrid models.}
    \label{tab:comparison_dgp}
    \begin{tabular}{lccc}
        \toprule
        Metric & Pure ML & Biasi & Bowring \\
         & DGP & Hybrid & Hybrid \\
        \midrule
        \multicolumn{4}{c}{\textbf{$\mathbf{7{,}350}$ Training Points ($\mathbf{80}$\% Case)}} \\
        \midrule
        $\upmu_\text{error}$ & $6.348\%$ & \cellcolor{gray!25}$4.196\%$ & $5.171\%$ \\ \hline
        $\text{Max}_{\text{error}}$ & $84.09\%$ & \cellcolor{gray!25}$37.29\%$ & $62.17\%$ \\ \hline
        $\text{Mean rStd}$ & $2.008\%$ & \cellcolor{gray!25}$1.793\%$ & $3.605\%$ \\ \hline
        $\text{Max rStd}$ & $28.63\%$ & \cellcolor{gray!25}$18.59\%$ & $61.63\%$ \\ \hline
        \text{rRMSE} & $9.413\%$ & \cellcolor{gray!25}$5.838\%$ & $7.680\%$ \\ \hline
        $F_{\text{error}}>10\%$ & $16.76\%$ & \cellcolor{gray!25}$7.835\%$ & $11.97\%$ \\ \hline
        $R^2$ & $0.9780$ & \cellcolor{gray!25}$0.9897$ & $0.9833$ \\
        \midrule
        \multicolumn{4}{c}{\textbf{$\mathbf{9}$ Training Points ($\mathbf{0.1}$\% Case)}} \\
        \midrule
        $\upmu_\text{error}$ & $62.92\%$ & $11.06\%$ & \cellcolor{gray!25}$10.32\%$ \\ \hline
        $\text{Max}_{\text{error}}$ & $881.5\%$ & $96.69\%$ & \cellcolor{gray!25}$88.76\%$ \\ \hline
        $\text{Mean rStd}$ & $9.839\%$ & $8.807\%$ & \cellcolor{gray!25}$7.362\%$ \\ \hline
        $\text{Max rStd}$ & \cellcolor{gray!25}$44.65\%$ & $681.9\%$ & $218.6\%$ \\ \hline
        \text{rRMSE} & $100.9\%$ & $15.74\%$ & \cellcolor{gray!25}$15.28\%$ \\ \hline
        $F_{\text{error}}>10\%$ & $87.49\%$ & $38.74\%$ & \cellcolor{gray!25}$35.26\%$ \\ \hline
        $R^2$ & $-0.0745$ & $0.9537$ & \cellcolor{gray!25}$0.9556$ \\
        \bottomrule
    \end{tabular}
\end{table}

The parity plots with uncertainty overlays were once again generated for the limited training dataset case. These plots are nearly identical in behavior to those of the BNN, showing a complete deterioration of the pure ML model's predictions with a similar relative standard deviation spread. Because of this redundancy, these parity plots are omitted from this section.

\begin{figure}[ht!]
    \centering
    \begin{subfigure}{0.49\textwidth}
        \centering
        \includegraphics[width=\linewidth]{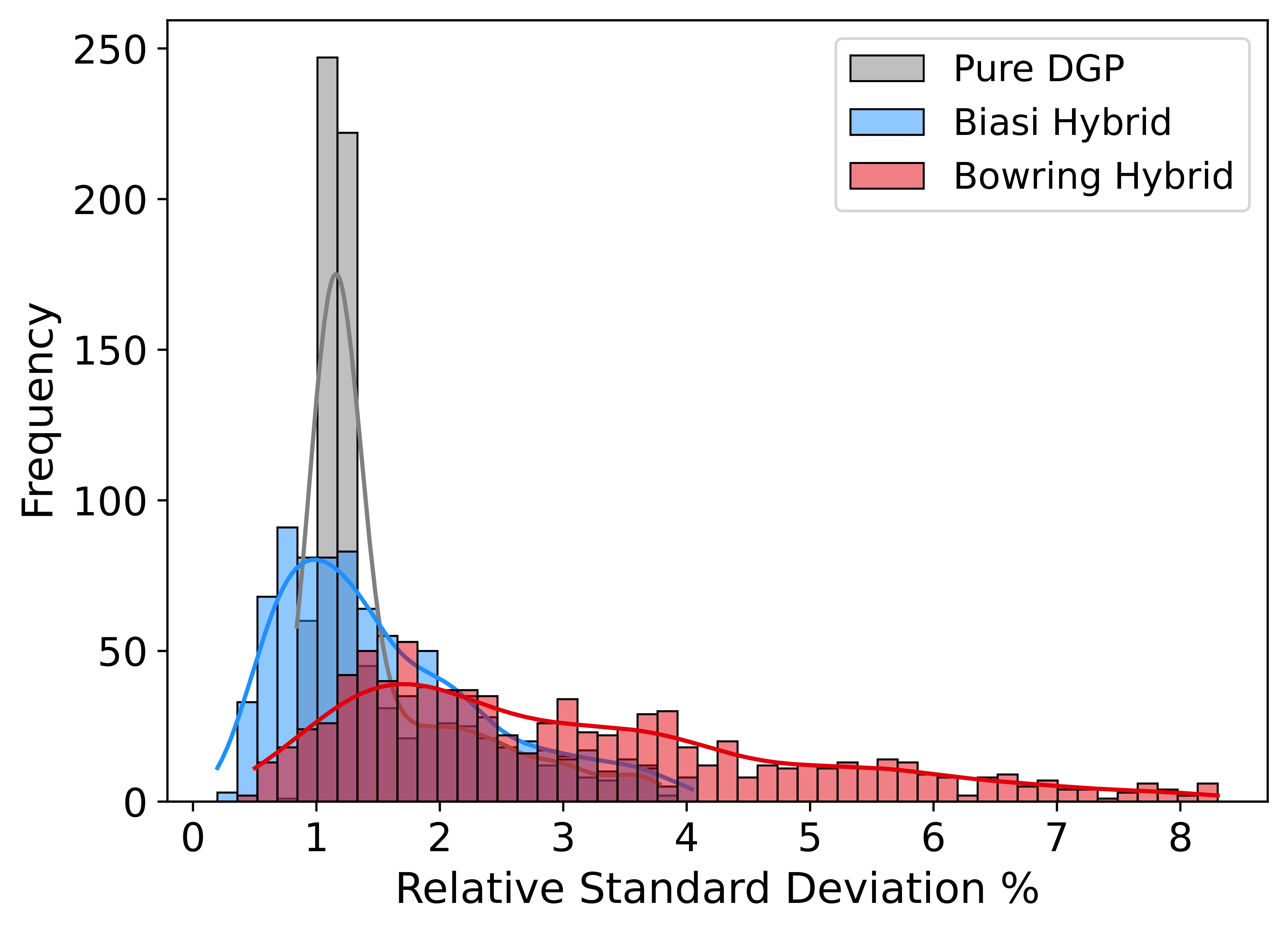}
        \caption{80\% Training Size}
    \end{subfigure}
    \begin{subfigure}{0.49\textwidth}
        \centering
        \includegraphics[width=\linewidth]{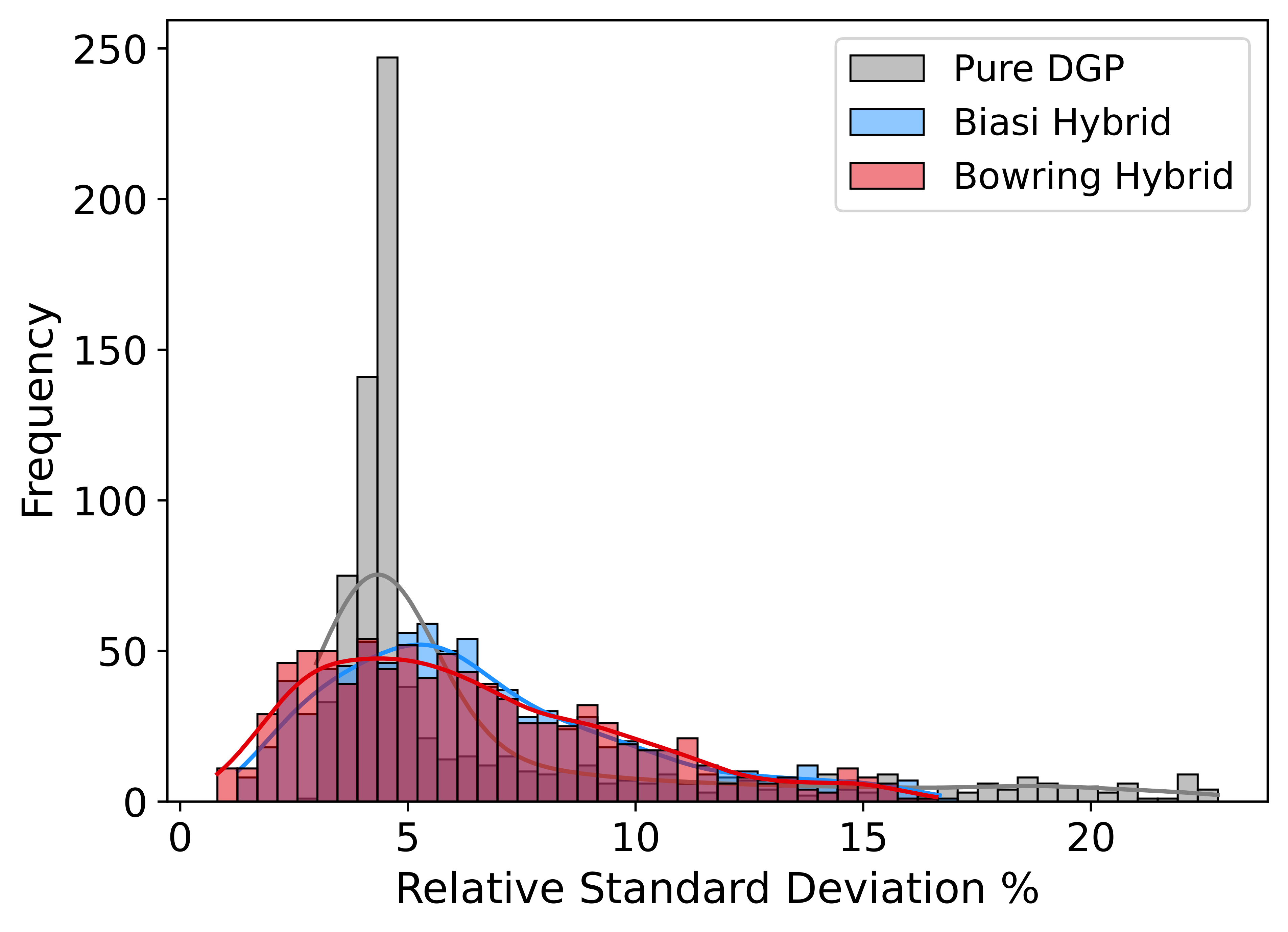}
        \caption{0.1\% Training Size}
    \end{subfigure}
    \caption{Distributions of the rStds for the DGP approach: histograms and KDE curves. Outlier values are removed to improve readability.}
    \label{fig:rstd_dist_dgp}
\end{figure}

The relative standard deviations for each of the points were extracted and plotted to assess the behavior of this distribution, as shown in Figure \ref{fig:rstd_dist_dgp}. Because of the large maximum rStd values in the hybrid models (681.9\% for Biasi and 218.6\% for Bowring in the limited training data case), all outliers are removed from Figure \ref{fig:rstd_dist_dgp} for visual purposes. In the plentiful training dataset case, the pure ML model has the smallest spread about its mean rStd value of 2.008\%. Although the Biasi hybrid model shows a smaller mean relative standard deviation value compared with that of the pure ML model, its standard deviation for this distribution is larger. The Bowring hybrid model has the largest mean, standard deviation, and maximum. These models' trends are similar in the limited training data case but with larger means, maximums, and standard deviations.

\begin{figure}[ht!]
    \centering
    \begin{subfigure}{0.49\textwidth}
        \centering
        \includegraphics[width=\linewidth]{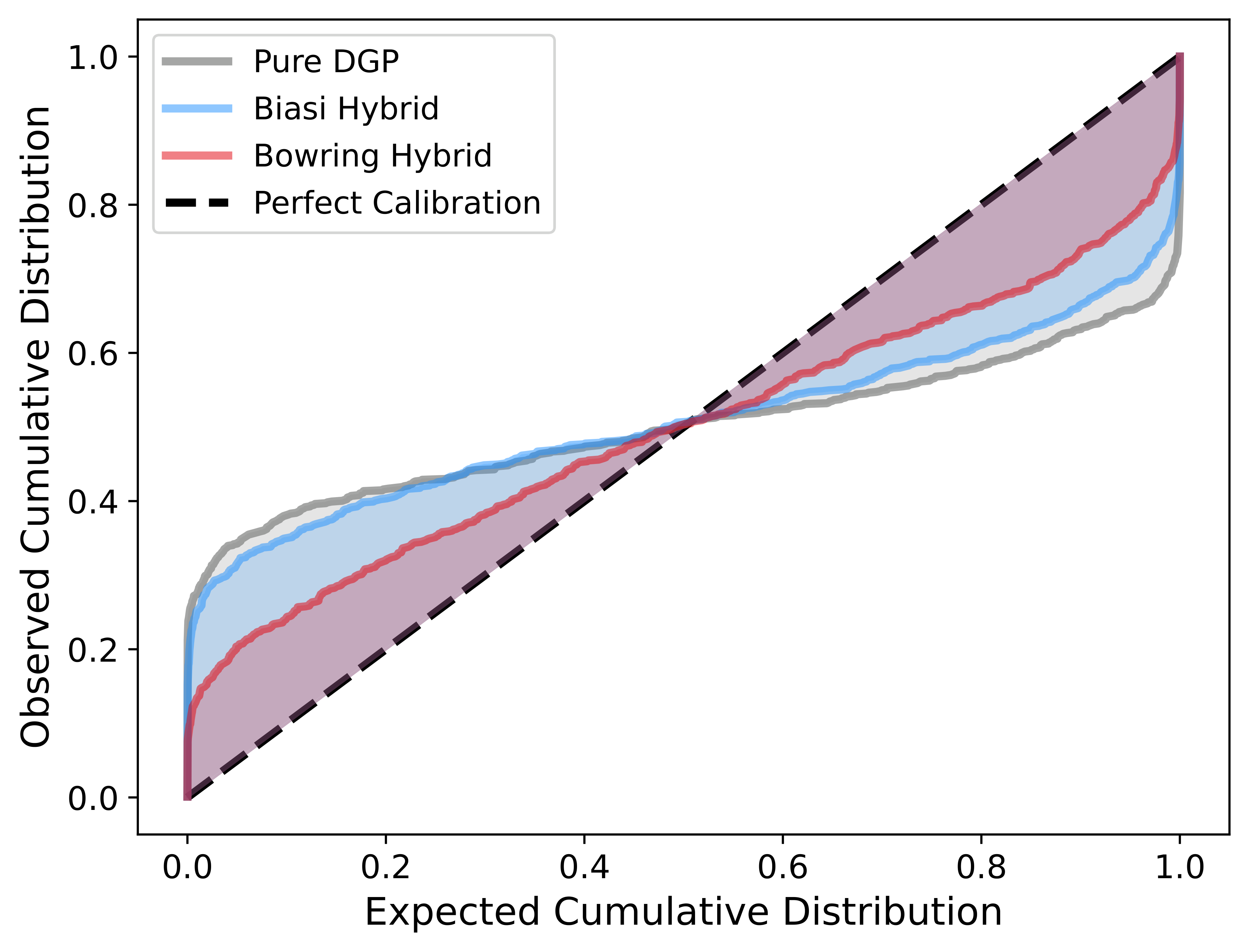}
        \caption{80\% Training Size}
    \end{subfigure}
    \begin{subfigure}{0.49\textwidth}
        \centering
        \includegraphics[width=\linewidth]{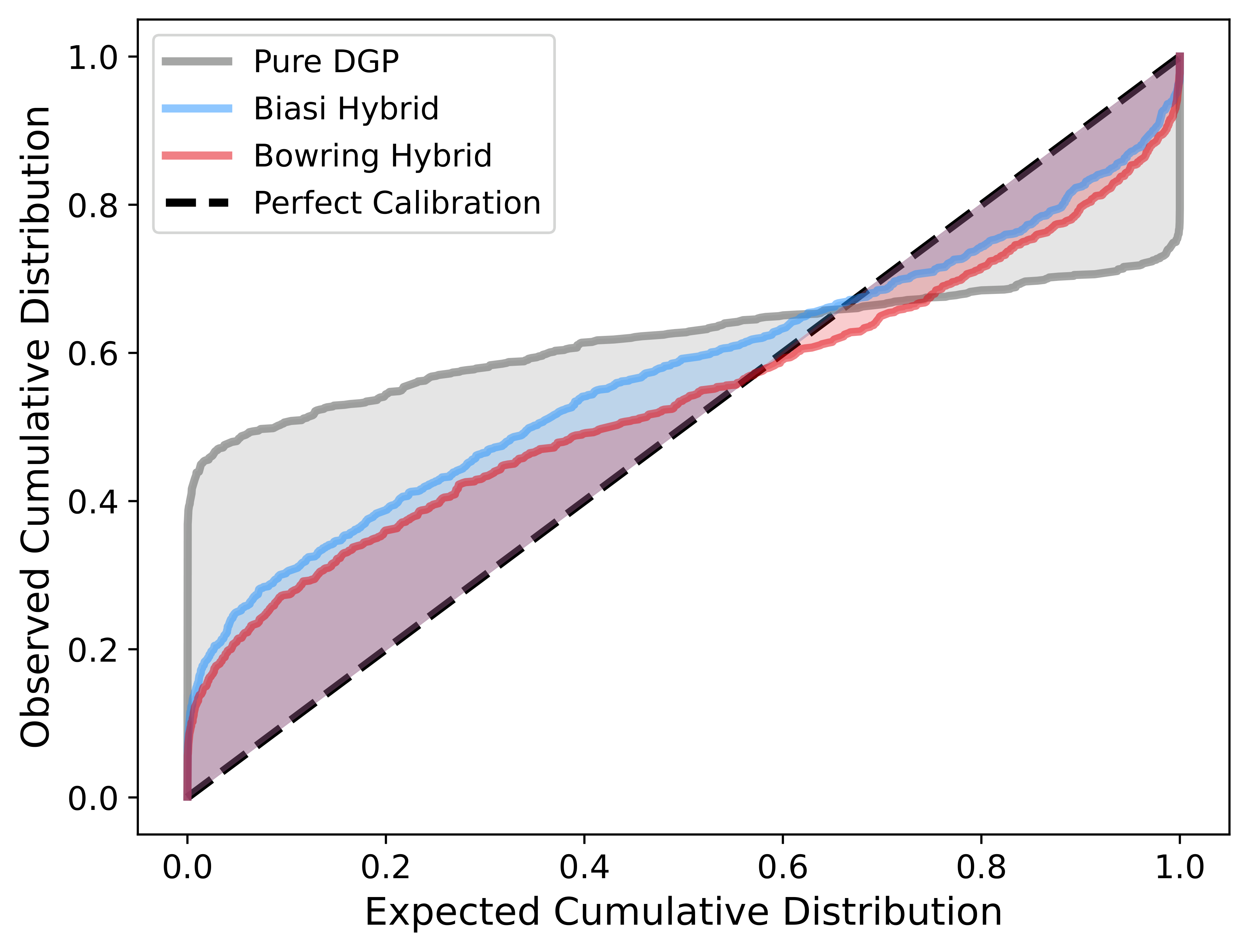}
        \caption{0.1\% Training Size}
    \end{subfigure}
    \caption{Calibration curves for the DGP approach.}
    \label{fig:calibration_dgp}
\end{figure}

The UQ calibration curves were then constructed. These curves are provided in Figure \ref{fig:calibration_dgp}. The Bowring hybrid model in both training data scenarios is the best calibrated compared with the other models, with miscalibration areas of 0.1063 in the plentiful case and 0.0944 in the limited case. All six of the cases between the two data scenarios have overconfident uncertainty estimates, leading to smaller-than-expected prediction intervals. By contrast, the pure ML models are the most poorly calibrated of the six DGP models (miscalibration areas of 0.1737 and 0.2085 for the plentiful and limited training scenarios, respectively). The Biasi hybrid model falls in between and also shows better calibration in the 0.1\% scenario (miscalibration area of 0.1115) than in the 80\% training scenario (miscalibration area of 0.1556).

\subsection{Direct Comparison between Methods}

\begin{figure}[ht!]
    \centering
    \includegraphics[width=0.5\linewidth]{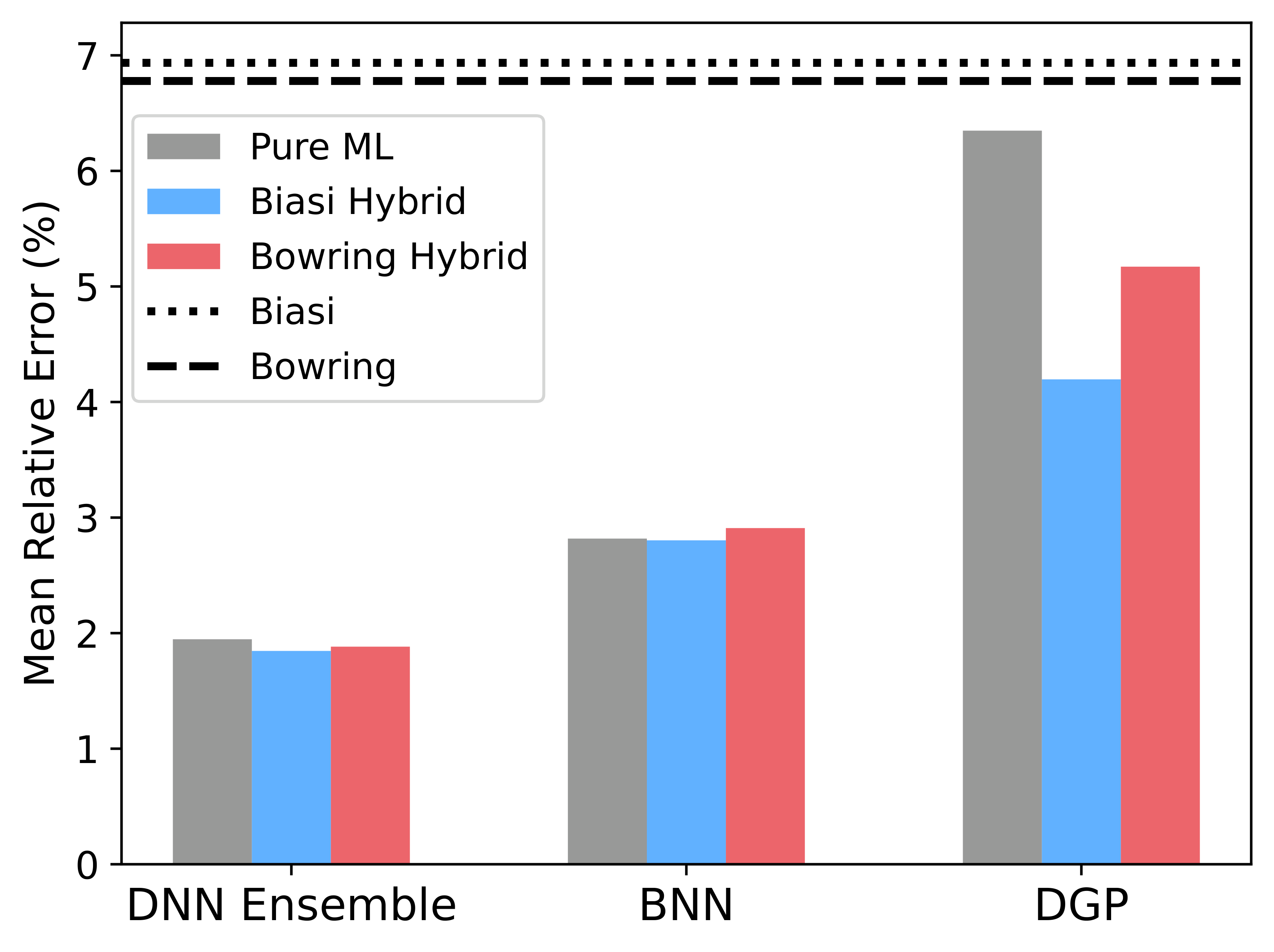}
    \caption{Comparison of $\upmu_{\text{error}}$ in the 80\% training size case.}
    \label{fig:mu_comparison}
\end{figure}

With the results above, direct comparisons can be made across the ML techniques (DNN ensemble, BNN, DGP) rather than comparisons simply being made between the pure and hybrid approaches performed within each. The first obvious metric to compare is the mean relative error. These values for the nine models of the 80\% training set size case are visually compared in Figure \ref{fig:mu_comparison}. Every model of the BNN and DGP has an increased error value compared with that of the ensemble; the DGP has the largest error of the three methods (6.348\%). The DNN ensemble and BNN show relatively even performance between pure and hybrid models in this training size case, with the pure DGP model having error values a full percentage point above the closest hybrid model (6.348\% compared with the Bowring hybrid's 5.171\%). The DGP's pure ML model has the largest relative error of any model (over 4 percentage points above any DNN ensemble model), and the hybrids follow. All nine models, however, have a smaller error than when the bare Biasi and Bowring correlations are applied to the dataset. The DNN ensemble has the smallest error overall and consequently the largest difference compared with that of the bare correlations; its relative error values are below 2\% (compared with the correlations' 6.935\% and 6.778\%). 

\begin{figure}[ht!]
    \centering
    \includegraphics[width=\linewidth]{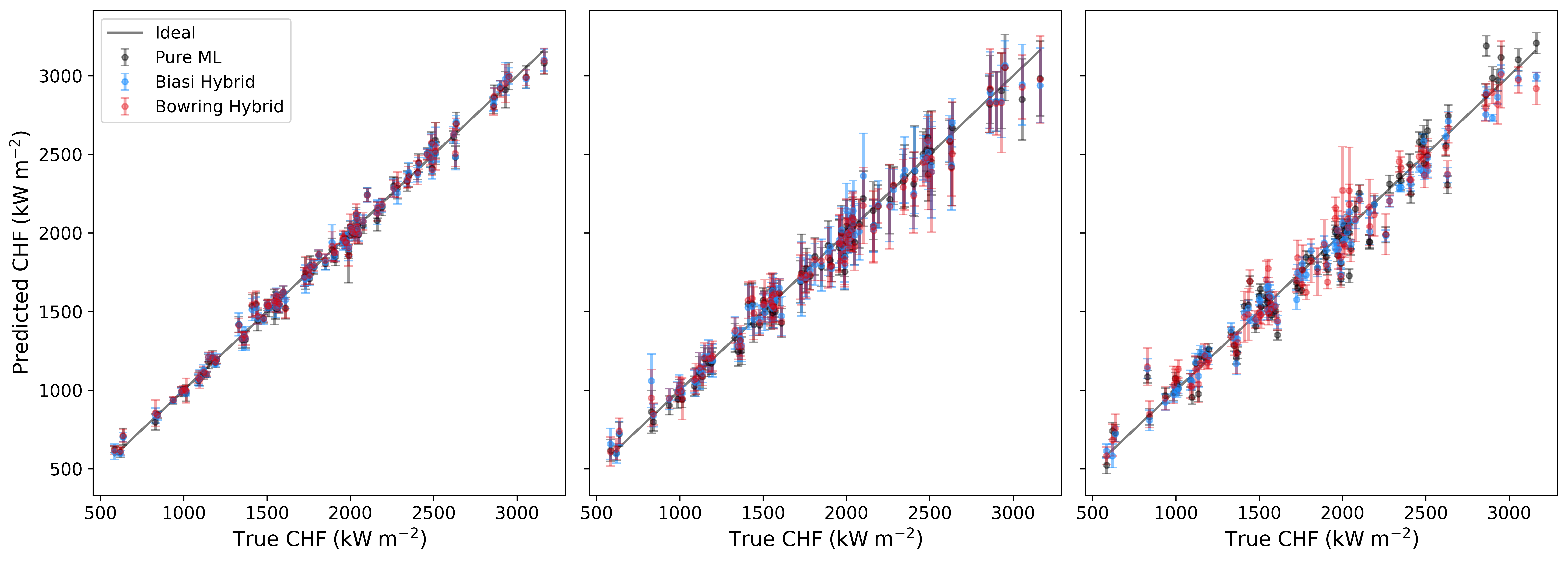}
    \begin{subfigure}[b]{0.32\textwidth}
        \caption{DNN Ensemble}
    \end{subfigure}
    \begin{subfigure}[b]{0.32\textwidth}
        \caption{BNN}
    \end{subfigure}
    \begin{subfigure}[b]{0.32\textwidth}
        \caption{DGP}
    \end{subfigure}
    \caption{Parity plots for BWR-range pressures comparing pure ML, Biasi hybrid, and Bowring hybrid across ML methods with $2\upsigma$ error bars in the 80\% training size case.}
    \label{fig:parity_2sigma_80p}
\end{figure}

The test set entries were then filtered using a ``BWR'' pressure range, set between 6,900~\si{\kilo\pascal} and 7,200 \si{\kilo\pascal} \cite{todreas2021nuclear}, to reduce the number of points for graphical analysis. This process was used for all the 80\% and 0.1\% training size cases. The remaining values were then plotted with prediction intervals created from their $2\upsigma$ values to assess the uncertainty distribution while also considering parity. In a normal distribution, $2\upsigma$ encloses roughly 95.4\% of all values. Prediction intervals in this case are an estimate of the interval in which a new prediction of a given input combination will fall inside with a specified probability. Notably, these sigma values are of the same units of CHF and are not relative values as previously discussed. The first, Figure \ref{fig:parity_2sigma_80p}, contains values for all models trained with the 80\% size case. An initial visual inspection indicates a clear difference between the prediction intervals of the DNN ensemble and those of both the BNN and DGP methods in all cases. The DNN ensemble's bars are relatively small along the identity line, with gradually increasing magnitudes with increasing CHF and a few larger values in the central region. Additionally, there is a tight adherence of mean values to this line, as previously indicated by the high overall accuracy. Each of the three models of the DNN ensemble indicates comparable standard deviation values.

Both the BNN and DGP models have relatively higher amounts of variance in each of the points' prediction intervals, particularly of note in all cases of the BNN and in the Bowring model of the DGP hybrid cases. All three models of the BNN have prediction intervals of similar magnitude, with a general trend of increasing height as CHF increases. This nonuniform distribution in uncertainty is heteroscedastic and can be due to multiple factors, including data variability and potential systematic biases in the model's weights. Notably, the scales of both the \textit{x}-axes and \textit{y}-axes are different from those of the previous parity plots, making the $2\upsigma$ values seem larger in comparison. Although the DGP models have larger error values compared with those of the BNN models, the DGP models' reported uncertainties are smaller in magnitude, with the largest located around the central region of the filtered CHF values. 

\begin{figure}[ht!]
    \centering
    \includegraphics[width=\linewidth]{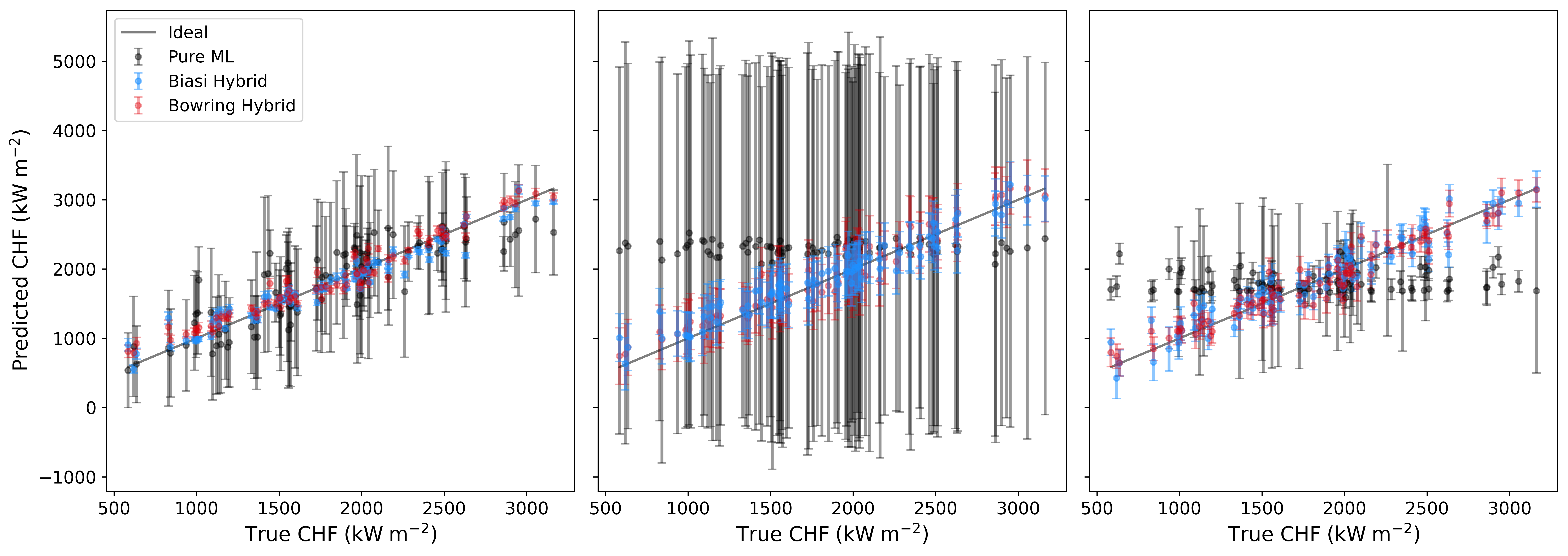}
    \begin{subfigure}[b]{0.32\textwidth}
        \caption{DNN Ensemble}
    \end{subfigure}
    \begin{subfigure}[b]{0.32\textwidth}
        \caption{BNN}
        \label{subfig:parity_2sigma_001p_bnn}
    \end{subfigure}
    \begin{subfigure}[b]{0.32\textwidth}
        \caption{DGP}
        \label{subfig:parity_2sigma_001p_dgp}
    \end{subfigure}
    \caption{Parity plots for BWR-range pressures comparing pure ML, Biasi hybrid, and Bowring hybrid across ML methods with $2\upsigma$ error bars in the 0.1\% training size case.}
    \label{fig:parity_2sigma_001p}
\end{figure}

The same process of constructing parity plots with prediction intervals was used for all models in the limited data scenario, which is shown in Figure \ref{fig:parity_2sigma_001p}. Each of the figures contains significantly different prediction interval behavior compared with those of Figure \ref{fig:parity_2sigma_80p}. The purely data-driven ML models have intervals with widths several times those of the hybrid models. This is particularly of note in Figure \ref{subfig:parity_2sigma_001p_bnn}, which corresponds to the BNN cases. Every prediction made by the pure ML model is shown with $2\upsigma$ widths on the order of 5000 \si{\kilo\watt\per\square\meter}, and all means are clustered about a single predicted value around 2000 \si{\kilo\watt\per\square\meter}. Both hybrid cases do not exhibit this behavior but rather follow the expected trend with means about the identity line and more reasonable uncertainties. The same effect is shown in Figure \ref{subfig:parity_2sigma_001p_dgp} but with pure DGP uncertainties fluctuating greatly, even though the means predict about the same value. These behaviors, in addition to the increased error and uncertainties shown in the DNN ensemble's parity plot, support the conclusion that these pure ML approaches offer very little usefulness in cases of scarce training data. Notably, these uncertainty estimates are not meant to be compared between each method (DNN ensemble, BNN, DGP) because they are not well calibrated (as discussed in the previous subsections) and would not be deployed for real-world use.

Each of the three methods was assessed using three different analysis techniques: performance metrics/parity, uncertainty distribution, and calibrations. To implement one of these approaches in a production code, a prospective model must be highly accurate and interpretable and have small uncertainty estimates that are well calibrated. With these criteria, the DGP approaches can be quickly eliminated on the basis of their relatively high error (4.196\% in the case of the Biasi hybrid), in addition to their uncalibrated uncertainty behavior (despite small reported uncertainties). The BNN models, though well calibrated in UQ, have larger reported uncertainties, along with greater error metrics, in comparison with the DNN ensemble. The DNN ensemble in all three cases (pure ML and the correlation hybrids) obtained the most favorable performance metrics, in addition to small, well-calibrated uncertainties with a significantly tighter distribution compared with those of the BNN models. Although the DNN ensembles were shown to be slightly less calibrated with regard to the BNN models, they are still considered to be well calibrated. Of the DNN ensemble models, the hybrids are structurally the most interpretable, which is the final criterion for model selection. Because of these factors, the DNN-based Biasi hybrid model is a competitive candidate for future deployment in a production code.

\section{Conclusions}\label{sec:conclusion}

This study developed and validated an uncertainty-aware hybrid modeling approach that combines ML with physics-based models to improve the prediction of CHF in nuclear reactors. The use of ML techniques, paired with empirical correlations, enables the hybrid model to capture underlying relationships in the correlations' residuals computed using measured data. 
This approach reduces the potential for non-physical outputs and increases interpretability as the output is partially constrained by the domain knowledge-based model. For all aspects of this study, the public CHF dataset used to construct the 2006 Groeneveld LUT was used after filtering to DO conditions as well as the correlations' ranges. The Biasi and Bowring correlations were used as the base models for the hybrid models with three different ML UQ techniques: DNN ensembles, BNNs, and DGPs. A pure ML model configuration without a base model was also considered as a baseline. For each of these combinations, a plentiful training dataset size (7,350 training points) and a limited scenario (9 training points) were considered to assess the hybrid models' resistance to data scarcity.

Performance and uncertainty of the trained models' predictions were quantified with seven statistical measures and three analysis techniques: (1) parity plots to inspect accuracy, (2) relative standard deviation distributions to assess the behavior of uncertainty, and (3) calibration curves to ensure calibration of the reported uncertainties. This process not only identifies the models with the most favorable error performance but also identifies those with the smallest uncertainty while ensuring the quality of those uncertainty estimates.

The Biasi hybrid DNN ensemble was determined to have the most favorable behavior in most measures, particularly in the plentiful training data scenario. In the limited data scenario, this model outperformed the stand-alone Biasi correlation, even with only nine training points. The BNN-based models have slightly increased error and uncertainty metrics compared with those of the DNN ensemble, but with the best UQ calibration in the plentiful data case. The DGP-based models have the poorest performance in nearly all measures, with large UQ miscalibration errors. Both the pure ML BNN and DGP models broke down in the limited data scenario, predicting nearly the same value for all input combinations. In all cases, the hybrid models outperformed the pure ML configurations.

The results of this study indicate that the use of knowledge-based hybrid ML modeling approaches to predict CHF values offers enhanced accuracy, interpretability, and a built-in resistance to performance degradation resulting from data scarcity. Implementation of these models within a production application, such as the CTF thermal-hydraulics code, would further demonstrate the feasibility of using these methods outside a pure research setting and provide a direct pathway for use in industrial applications. Other future work will include additional refinement and validation with a set of more complicated tests. The use of transfer learning may also be explored to extend this method in the case of rod bundles instead of single channels.

\section*{Acknowledgments}

This research was supported in part by an appointment to the US Department of Energy's Omni Technology Alliance Internship Program, sponsored by DOE and administered by the Oak Ridge Institute for Science and Education (ORISE). The authors from North Carolina State University were also funded by the DOE Office of Nuclear Energy Distinguished Early Career Program (DECP) under award number DE-NE0009467.

\bibliography{bibliography.bib}

\end{document}